\documentclass[aoas]{imsart}
\doi{10.1214/22-AOAS1705}
\copyrightowner{Institute of Mathematical Statistics}

\RequirePackage{amsthm,amsmath,amsfonts,amssymb}
\RequirePackage[authoryear]{natbib}
\RequirePackage[colorlinks,citecolor=blue,urlcolor=blue]{hyperref}
\RequirePackage{graphicx}
\RequirePackage{hhline,bm,xcolor,amsmath}
\RequirePackage{graphicx,multirow,booktabs,subfigure,comment}
\RequirePackage[linesnumbered,ruled]{algorithm2e}
\RequirePackage{caption}
\RequirePackage{mathtools}
\def\sign{\mathop{\rm Sign}}
\newcommand*{\EOP}{\hfill\ensuremath{\square}}

\newcommand{\CAE}{\text{CAE}}

\newcommand{\pkg}[1]{{\fontseries{m}\fontseries{b}\selectfont #1}}

\startlocaldefs
\theoremstyle{plain}

\newtheorem{theorem}{Theorem}[section]
\newtheorem{corollary}[theorem]{Corollary}

\newtheorem{lemma}[theorem]{Lemma}

\def\argmin{\mathop{\rm argmin}}

\DeclarePairedDelimiter\floor{\lfloor}{\rfloor}
\makeatletter
\newcommand*{\rom}[1]{\expandafter\@slowromancap\romannumeral #1@}
\theoremstyle{remark}
\newtheorem{definition}[theorem]{Definition}

\endlocaldefs

\begin{document}
\begin{frontmatter}
\title{Data-Adaptive Discriminative Feature Localization with Statistically Guaranteed Interpretation}
\runtitle{Data-Adaptive Discriminative Localization}

\begin{aug}
 

\author{\fnms{Ben} \snm{Dai}},
\author{\fnms{Xiaotong} \snm{Shen}},
\author{\fnms{Lin Yee} \snm{Chen}},
\author{\fnms{Chunlin} \snm{Li}}
\and
\author{\fnms{Wei} \snm{Pan}}


\address{\normalsize{The Chinese University of Hong Kong and University of Minnesota}}




\end{aug}

\received{\smonth{3} \syear{2022}}
\revised{\smonth{8} \syear{2022}}

\begin{abstract}

In explainable artificial intelligence, discriminative feature localization is critical to reveal a blackbox model's decision-making process from raw data to prediction. 
In this article, we use two real datasets, the MNIST handwritten digits and MIT-BIH Electrocardiogram (ECG) signals, to motivate key characteristics of discriminative features, namely \textit{adaptiveness}, \textit{predictive importance} and \textit{effectiveness}. 
Then, we develop a localization framework based on adversarial attacks to effectively localize discriminative features. In contrast to existing heuristic methods, we also provide a statistically guaranteed interpretability of the localized features by measuring a generalized partial $R^2$. 
We apply the proposed method to the MNIST dataset and the MIT-BIH dataset with a convolutional auto-encoder. In the first, the compact image regions localized by the proposed method are visually appealing. Similarly, in the second, the identified ECG features are biologically plausible and consistent with cardiac electrophysiological principles while locating subtle anomalies in a QRS complex that may not be discernible by the naked eye. Overall, the proposed method compares favorably with state-of-the-art competitors. Accompanying this paper is a Python library \pkg{dnn-locate} that implements the proposed approach.

\end{abstract}


\end{frontmatter}

\section{Introduction}
\label{sec:intro}
The empirical success of machine learning in real applications has profound impacts on many scientific and engineering areas, including image analysis \citep{lecun1989backpropagation, he2016deep}, recommender systems \citep{wang2015collaborative}, natural language processing \citep{hochreiter1997long}, drug discovery \citep{vamathevan2019applications}, protein structure prediction \citep{jumper2021highly, evans2021protein}.
However, the nature of a black-box model makes it challenging to interpret its
decision-making process. The lack of interpretability hinders transparency, trust, and understanding of scientific discovery. To meet challenges, Explainable AI (XAI) is emerging, which includes localizing discriminative features attributing to a model's predictive performance, shaping or confirming human intuitions and knowledge, for instance, visual explanation on image recognition.

\subsection{Motivation: DL discriminative localization in the MIT-BIH ECG dataset}
\begin{figure}[h]
  \centering
    \includegraphics[scale=.28]{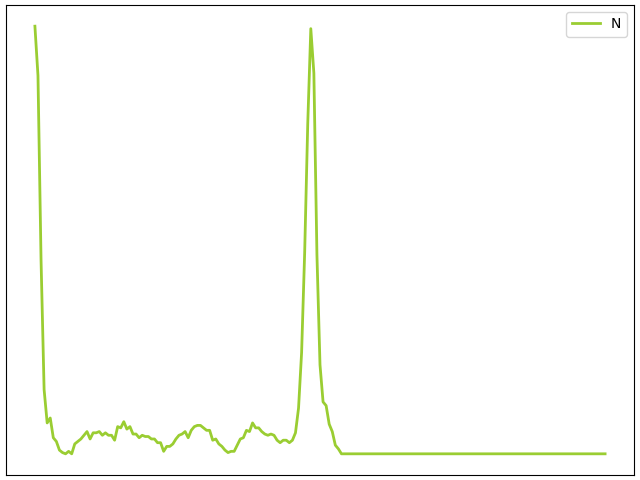}
    \includegraphics[scale=.28]{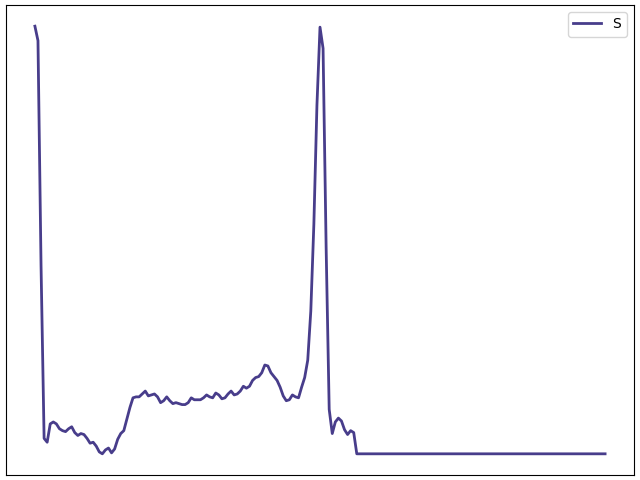}
    \includegraphics[scale=.28]{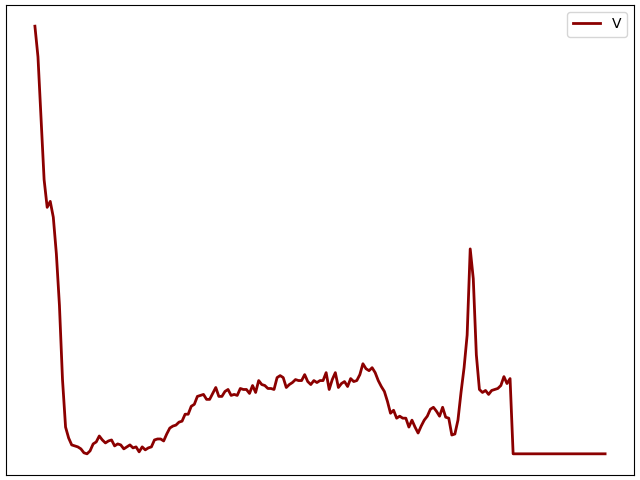}
    \includegraphics[scale=.28]{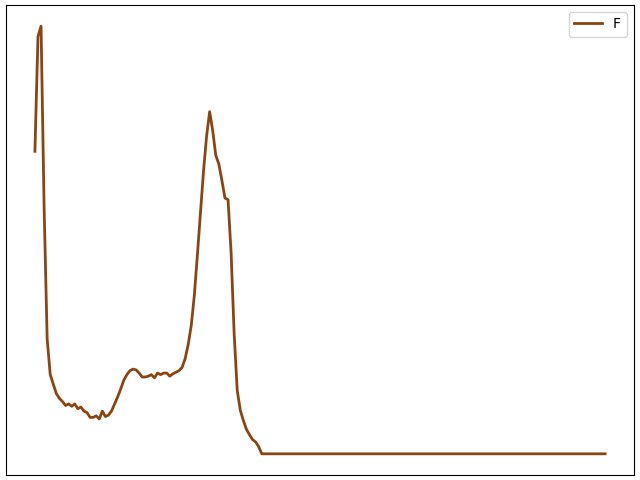}
    \includegraphics[scale=.28]{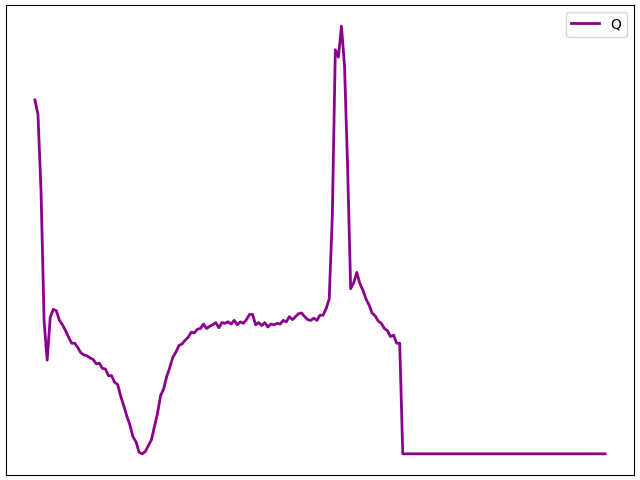}
\caption{Five classes of ECG beat: \{`N': normal, left/right bundle branch block, atrial escape, nodal escape\}, \{`S': atrial premature, aberrant atrial premature, nodal premature, supra-ventricular premature\}, \{`V': premature ventricular contraction, ventricular escape\}, \{`F': fusion of ventricular and normal\}, \{`Q': paced, fusion of paced and normal, unclassifiable\}. }
\label{fig:class}
\end{figure}

Our investigation responds to the need for locating features that are most critical to a learning outcome. The present study is motivated by the MIT-BIH ECG dataset and the MNIST dataset. 
Specifically, the MNIST dataset serves as a benchmark for studying XAI methods \citep{ribeiro2016should,lundberg2017unified}, in part because the results of Localization could be easily evaluated by human intuition. 
As demonstrated in Figures \ref{fig:demo_motivation} and \ref{fig:demo}, localized image pixels explain how a deep convolutional network  differentiates digits `7' and `9' on the MNIST data.
A more substantial medical application is based on the MIT-BIH ECG dataset, this dataset is a commonly used ECG benchmark dataset, which consists of ECG recordings from 47 different subjects recorded at the sampling rate of 360Hz by the BIH Arrhythmia Laboratory. 
Each beat is annotated into 5 different classes under the Association for the Advancement of Medical Instrumentation (AAMI) EC57 standard \citep{stergiou2018universal}: 'N', 'S', 'V', 'F', and 'Q'. One random sample per class is demonstrated in Figure \ref{fig:class}.

\begin{figure}[h]
  \centering
    \includegraphics[scale=.24]{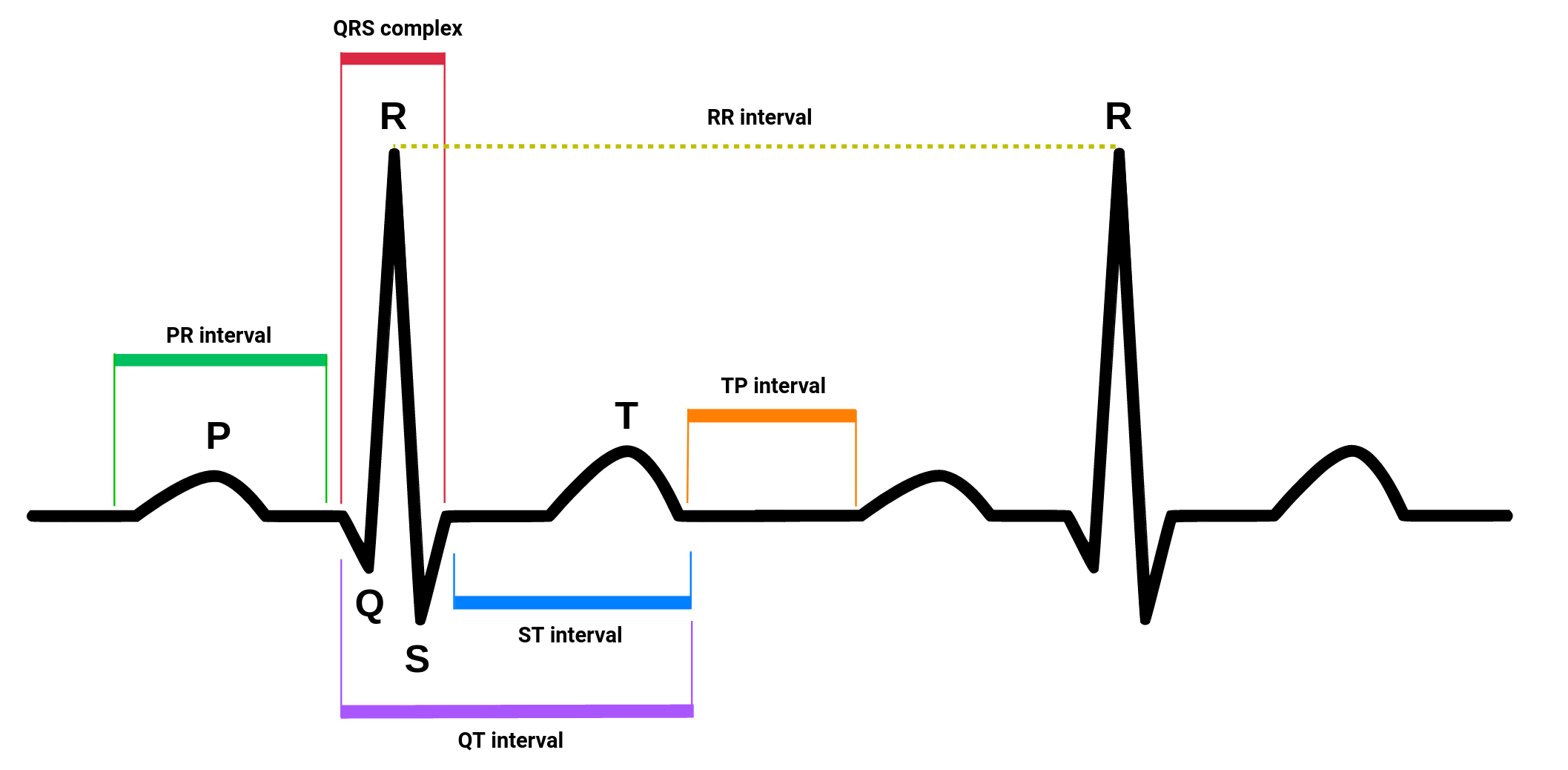}
\caption{A typical ECG signal with its most common waveforms, where important points and intervals are marked. 
}
\label{fig:QRS}
\end{figure}

Broadly speaking, The existing ECG diagnosis methods in the literature can be categorized into two: conventional machine learning (ML) and deep learning (DL) methods.
Conventional ML methods first extract manually-crafted features based on ECG background knowledge and some signal morphological technique, including the QRS complex, T wave, R-R interval, S-T interval \citep{wasimuddin2020stages}; see Figure \ref{fig:QRS}.
Next, conventional classification methods, such as support vector machines (SVMs; \cite{cortes1995support}), random forest \citep{breiman2001random}, and gradient boosting \citep{friedman2001greedy}, can be used to implement ECG diagnosis under a supervised learning framework based on extracted features \citep{jambukia2015classification}. 
However, conventional methods strongly depend on the quality of the manually-defined features, which are limited by existing domain knowledge. Specifically, the manually-crafted morphological features may not be able to capture all predictive information in the original ECG signals \citep{bharti2021prediction,thygesen2007universal}. 
Moreover, it is also challenging to perfectly extract morphological features from ECG signals due to electrical noise caused by tray magnetic fields and accessories that vibrate \citep{elgendi2013fast}. Therefore, certain biases may be introduced during feature engineering, thus hampering the accuracy of ECG diagnosis.


Recently, deep learning has garnered considerable success in ECG diagnosis. DL differs from conventional ML methods in directly fitting a neural network based on raw ECG signals without feature engineering to extract manual-crafted features. DL models have recently delivered superior performance in the classification of ECG diagnosis.
For instance, existing convolutional neural networks \citep{attia2019artificial,rajpurkar2017cardiologist,ko2020detection} achieved over 93\% heartbeat classification accuracy. In contrast to conventional ML methods, DL models can effectively and adaptively extract the underlying information from raw data.
Alternatively, the DL models may localize some novel discriminative features that even ECG experts may not be aware of nor can discern. However, despite their merits, DL models are often referred to as a blackbox, referring to the seeming mystery of their decision-making processes. The lack of interpretable features relevant to the prediction  stands out as a significant barrier to the clinical use of their routine.
Therefore, our primary goal is to develop a localization framework to unmask unknown discriminative features of blackbox models to help bridge the bench-to-bedside gap and explore the domain knowledge of interpreting ECGs.

\begin{figure}[h]
  \centering
    \includegraphics[width=1.\textwidth]{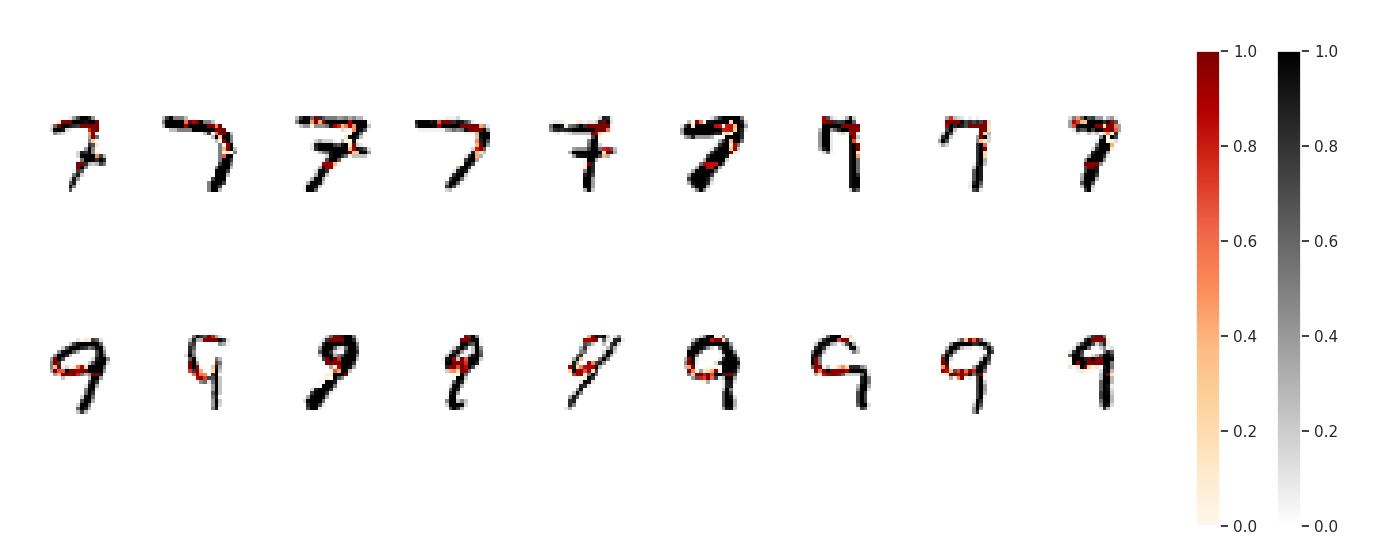}
   \caption{Data-adaptive localized discriminative features (red) for the proposed method with $\tau = 20$ based on different `7' and `9' digits (black). The gray color-bar indicates gray scale of original images, and the red color-bar indicates the proportion.}
   \label{fig:diff_demo}
\end{figure}

Discriminative feature localization for DL models is important but challenging. The major difficulties include (i) discriminative features are {data-dependent} on an input instance. For example, in the MNIST or ECG dataset, the location of discriminative features may differ with inputs; see Figures \ref{fig:diff_demo} and \ref{fig:ECG}. On this premise, classical variable selection methods based on tabular data are unsuitable without modification; instead, it requires \textit{data-adaptive} feature selection. (ii) A reliable statistical measure supported by theory is required to quantify \textit{predictive importance} of any discriminative feature. Most existing methods are heuristic and fail to interpret the localized features. (iii) As indicated in Figure \ref{fig:demo_motivation}, the localized features should effectively explain the discrimination of different outcomes. Hence, \textit{effectiveness} and \textit{predictive importance} should be simultaneously considered for selecting sensible discriminative features.




\subsection{Prior work and our contributions}


Three major approaches have emerged for discriminative feature localization, including two-stage methods, feature-importance-based methods, and backtracking methods. Specifically, 
two-stage methods use a simple explainable model, such as a local linear model, to approximate a complex blackbox model, and then to extract discriminative features.  In particular, a method called local interpretable model-agnostic explanations (LIME) \citep{ribeiro2016should} approximates a classification model by a local sparse linear model based on a kernel smoother as in \cite{davis2011remarks}, then highlights those features with positive linear coefficients. Deep-Taylor \citep{montavon2017explaining} expands and decomposes a neural network output in terms of its input variables and generates a heatmap by back-propagating explanations from output to input. 
Feature-importance-based methods rank each feature's contribution by its importance based on an approximating model in a two-stage method. For example, SHAP (SHapley Additive exPlanations) \citep{lundberg2017unified} develops a kernel method integrating LIME with the SHAP-value as the kernel weights and feature importance to quantify the contribution of features in an approximating local linear model.
The backtracking methods map the activation layers of a neural network back to the input feature space, identifying which input patterns contribute more to prediction.  In particular, \cite{zhou2016learning} uses the global average pooling (GAP) together with class activation mapping (CAM) at the last layer of a convolutional neural network (CNN). Then it
backtracks discriminative regions at the previous convolutional layers to the predicted scores. Gradient-CAM \citep{selvaraju2017grad} extends GAP to a general CNN model by computing the gradient of a decision score concerning the feature activation maps of a convolutional layer. Deconvnet \citep{zeiler2014visualizing} and Layer-wise Relevance Propagation (LRP) \citep{bach2015pixel} perform backtracking with a deconvolution and conservative relevance redistribution, respectively. Finally, Patternnet \citep{kindermans2017learning} identifies discriminative features by localizing the signal and noise directions for each neuron of a neural network.

Despite their merits, issues remain. First, a two-stage approach does not directly interpret an original model since discriminative features are localized by a simple approximation. For example, discriminative features generated by a linear approximation model \citep{ribeiro2016should,lundberg2017unified} may be neither discriminative nor interpretable in the original model.
Second, most existing methods are heuristic. As argued in \cite{tjoa2019survey}, an intermediate backtracking process for GAP, Gradient-CAM and LRP are not amenable to scrutiny. Moreover, Deconvnet and LRP fail to produce a theoretically correct explanation even for a linear model \citep{kindermans2017learning}. Finally, the above methods usually provide a dense representation of discriminative features, as suggested in Figure \ref{fig:camp_competitor}, yielding less effective interpretation.


There are three key contributions of our work in this paper: 
\begin{itemize}
  \item We propose a generalized partial $R^2$ in Definition \ref{def:r_squared} to quantify the degree of \textit{predictive importance} of discriminative features so that they can be interpreted similarly as in classical statistical analysis.
  \item The proposed localization framework \eqref{eqn:emp_model} is able to simultaneously consider both \textit{predictive importance} and \textit{effectiveness}. Specifically, as illustrated in Figures \ref{fig:demo} and \ref{fig:ECG}, it provides a \textit{flexible} framework to localize discriminative features corresponding to a certain amount of accuracy, as measured by an $R^2$.
  \item Through numerical experiments in Section \ref{sec:application} (the MNIST dataset), the localized discriminative features not only confirm the visual intuition but also are more efficient than the other existing methods. The numerical experiments in Section \ref{sec:ECG} suggest that localized ECG features are biologically plausible and consistent with cardiac electrophysiological principles, while locating subtle anomalies in sinus rhythm that may not be discernible by the naked eyes.
\end{itemize}


\section{Generalized partial \texorpdfstring{$R^2$}{TEXT} for discriminative localization}
\label{sec:R2}
In this section, we introduce generalized partial ${R}^2$ to quantify the degree of \textit{predictive importance} of discriminative features.
\subsection{Motivation}
In a learning paradigm, a prediction function $d$ is trained to predict an outcome $Y$ for a given instance $\bm X$, where $\bm{X}=(X_1,\cdots,X_p)^{\intercal}$ is a $p$-dimensional continuous feature vector.
Without loss of generality, each feature component $X_j$ is rescaled to $[0,1]$. {For example, in the MNIST dataset, $\bm{X}$ is a gray-scale image, and $Y$ is its associated digit label \citep{lecun-mnisthandwrittendigit-2010}.} 
To assess the performance, a loss function $L(\cdot, \cdot)$ is used, 
such as the cross-entropy loss $L(d(\bm{X}), Y) = - \bm{1}^\intercal_Y \log \big( \text{softmax}\big(d(\bm{X})\big) \big)$, where $\bm{1}_Y$ is the one-hot encoding of $Y$ and $\text{softmax}(\bm z) = (\text{softmax}(\bm z)_1,\cdots, \text{softmax}(\bm z)_p)^{\intercal}$ with $\text{softmax}(\bm z)_i={\exp(z_i)}/{\sum_{j} \exp(z_j)}$. 

\begin{figure}[h]
  \centering
    \includegraphics[scale=.35]{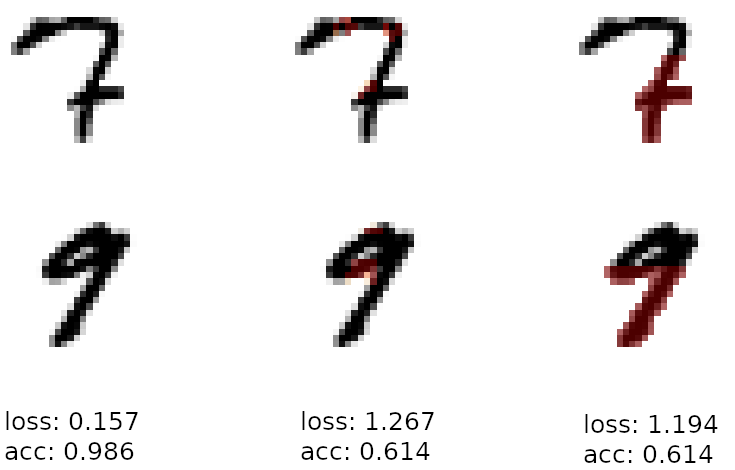}
    \rule[1ex]{\textwidth}{0.1pt}
    \includegraphics[scale=.33]{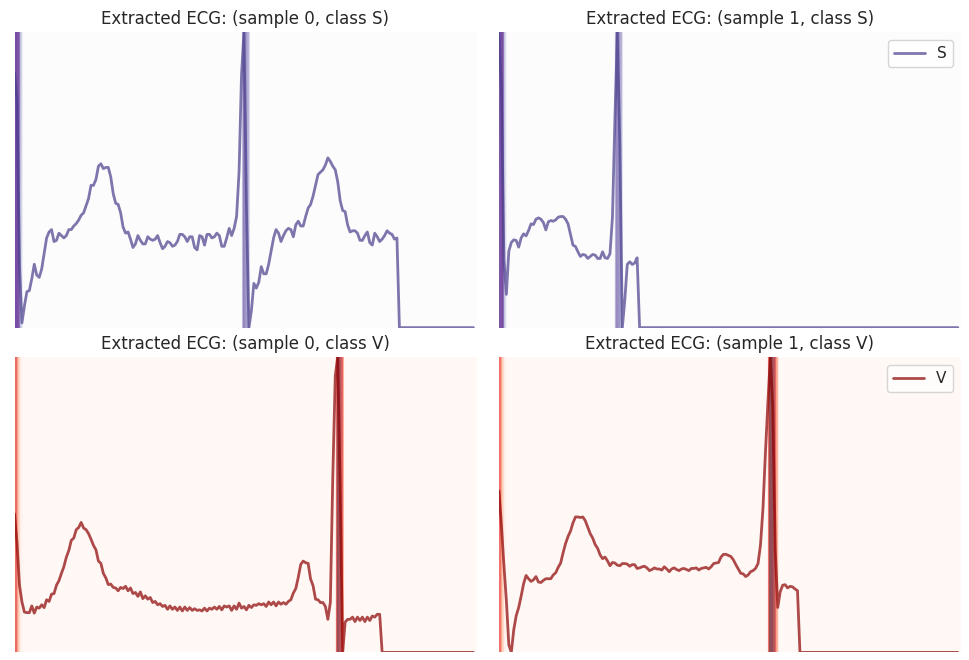}
\caption{Examples to illustrate the concepts of \textit{adaptiveness, predictive importance, and effectiveness} of discriminative features on the MNIST and MIT-BIH data. 
\textbf{Upper panel.} The left represents raw images of digits 7 and 9, the middle represents images with localized pixels (marked in red) by the proposed method, and the right represents images with localized pixels from row 15 to 28. 
Here `loss' and `acc' denote the cross-entropy and classification accuracy of a conventional neural network for each of the original and two disrupted dataset (by removing the localized regions of all images).
\textbf{Lower panel.} The top and bottom each show two extracted ECGs and their localized regions from class `S' (blue) or `V' (red) in the MIT-BIH data. Specifically, the red/blue solid lines are the extracted ECG signals, and the highlighted vertical blue/red bars are localized regions by the proposed method.
More discussion can be found in Sections \ref{sec:application} and \ref{sec:ECG}.}
\label{fig:demo_motivation}
\end{figure}

Our goal is to identify discriminative features that effectively disrupt or deteriorate the prediction performance of a given learner $d$. To proceed, we highlight three distinctive characteristics of discriminative features motivated from real applications, namely \textit{adaptiveness}, \textit{predictive importance}, and \textit{effectiveness}.
As an illustrative example based on the MNIST dataset, consider two localized feature sets in the left panel of Figure \ref{fig:demo_motivation}. The feature set removed in the middle or right panel decreases the predictive accuracy of $d$ by the same amount from 0.986 to 0.614, which suggests that the discriminate features should contribute largely to the predictive performance of $d$. Moreover, with the same amount of deterioration of performance, the highlighted features in the middle panel appear more compact, which we call more \textit{effective} in the sequel, and thus more preferred as discriminative features.
Furthermore, key characteristics are also captured by the MIH-BIH data. In particular, the amplitudes and locations of the QRS complexes \citep{kusumoto2020ecg}, as well as of P and T waves, varying across ECG signals even of the same class, dictate that the discriminative features should be \textit{adaptive} to the input ECG signals, as shown in the right panel of Figure \ref{fig:demo_motivation}. 
Note that the QRS complex corresponds to the spread of a stimulus through the ventricles and is usually the most visually important part of an ECG tracing \citep{kusumoto2020ecg}.
Moreover, ion channel aberrations and structural abnormalities in the ventricles can affect electrical conduction in the ventricles \citep{rudy2004ionic}, manifesting with \textit{subtle} anomalies in the QRS complex in sinus rhythm that may not be discernible by the naked eyes, yielding sparse or \textit{effective} discriminative features.

In summary, three distinctive characteristics of discriminative features are desired:
\begin{itemize}
\item \textit{Adaptiveness}. Discriminative feature extraction has to be \textit{adaptive} to an input instance and a specific learner $d$. For example, in the MNIST/MIT-BIH dataset, the location of discriminative features may differ with input images/signals.

\item \textit{Predictive importance}. The prediction accuracy of a learner $d$ would significantly deteriorate without discriminative features. Alternatively, discriminative features can explain a large proportion of its predictive performance.

\item \textit{Effectiveness}. Discriminative features should effectively describe the discrimination of the outcome. Therefore, under the same \textit{predictive importance}, the number/amount of localized discriminative features should be as small as possible. For example, compact localized pixels in the MNIST dataset or compact and accurate location of QRS complexes of ECG signals in the MIT-BIH ECG dataset.
\end{itemize}

To address \textit{adaptiveness}, we introduce a \textit{localizer} $\bm{\delta}(\bm{x})=(\delta_1(\bm{x}),\cdots,\delta_p(\bm{x}))^{\intercal}:\mathbb{R}^p \to \mathbb{R}^p$ to produce a disruption adaptively based on an instance $\bm x$ to yield disrupted features $\bm{x}_{\bm{\delta}} = \bm{x} - \bm{\delta}(\bm{x})$. Without loss of generality, assume that each $|\delta_j(\bm x)| \leq 1$ because $x_j$ is rescaled to be in $[0,1]$. In practice, the restriction $|\delta_j(\bm x)| \leq 1$ is usually met by construction, for example, in an auto-encoder in image classification, see Section \ref{sec:model_image} for illustration.

\subsection{Generalized partial \texorpdfstring{$R^2$}{TEXT}}

To measure the degree of \textit{predictive importance} of a localizer $\bm{\delta}(\cdot)$, we introduce a generalized partial $R^2$, which mimics the partial $R^2$ in regression \citep{nagelkerke1991note} and McFadden's $R^2$ \citep{mcfadden1973conditional} in classification. Specifically, the main idea of the partial $R^2$ is one minus the ratio of the full-model risk to the partial model risk. On this ground, we generalize the partial $R^2$ to blackbox models in Definition 1. 

\begin{definition}[Generalized partial $R^2$]
\label{def:r_squared}
Given a predictive model $d$, we define the generalized partial $R^2$ based on a localizer $\bm{\delta}(\cdot)$ as
\begin{align}
R^2(d, \bm{\delta}) = 1 - \frac{ \mathbb{E}\big( L ( d(\bm{X}), Y ) \big) }{ \mathbb{E}\big( L ( d(\bm{X}_{\bm{\delta}}), Y ) \big) }.
\end{align}
If $R^2(d, \bm{\delta}) \geq r^2$, we say that the localized features by $\bm{\delta}(\cdot)$ is $r^2$-discriminative. 
\end{definition}

The generalized partial $R^2$ is one minus the proportion of the risk on full features $\bm X$ over that of the disrupted features $\bm X_{\delta} = \bm{X} - \bm{\delta}(\bm{X})$. It is a natural and clear criterion to extend the classical $R^2$, and measure the \textit{predictive importance} of the features disrupted by a localizer. Specifically, a higher $R^2$ yields stronger \textit{predictive importance} of the localized discriminative features. When $\bm{\delta}(\bm{x})$ 
does not affect the performance of $d$, that is, $\mathbb{E}\big( L ( d(\bm{X}_{\bm{\delta}}), Y ) \big) = \mathbb{E}\big( L ( d(\bm{X}), Y ) \big)$, or $R^2(d, \bm{\delta}) = 0$, the localized features contain no information for prediction. 
On the other hand, $r_{\max}^2 = \max_{\bm{\delta}} R^2(d, \bm{\delta})$ 
the largest $R^2$ among all possible localizers, gives an upper bound of $R^2$. For instance, a localizer with each $\delta_j(\bm{x}) = \bm{x}_j$ disrupts extremely by removing all features, which forces a learner $d$ to predict without features. In general, $0 < R^2 < r_{\max}^2$ indicates the percentage of performance explained by $\bm{\delta}$.

\section{Methods} Our main idea of identifying effective discriminative features is to seek a localizer $\bm{\delta}(\bm{x})$ yielding the most effective disruption of the features to reduce the prediction accuracy of a learner $d$. 
\label{sec:method}
\subsection{A discriminative localization framework}
In Figure \ref{fig:demo_motivation}, the $r^2$-discriminative localizer in the right panel is ineffective, although it also affects the same amount of prediction accuracy. Therefore, discriminative features should have an \textit{effective} (or compact) representation, in addition to their contribution to a learner's prediction accuracy. 

To achieve this goal, we introduce an activity $L_1$-regularizer $J(\bm{\delta})$ to quantify the \textit{effectiveness} of a localizer,
\begin{equation}
\label{eqn:l1-norm}
J(\bm{\delta}) = \sup_{\bm{x}} \sum_{j=1}^p \big|\delta_j(\bm{x}) \big|.
\end{equation}
The benefits of this regularizer are two folds. First, it coincides with greedy feature selection results as indicated in Appendix A. Second, the supremum in \eqref{eqn:l1-norm} makes the localized features more balance over an entire
sample, as suggested in Section \ref{sec:application}. Moreover, we specify $\| \bm{\delta}(\bm{x}) \|_\infty \leq 1$ for any $\bm{x}$, to control the magnitude of the disruption. This requirement can be trivially satisfied, for instance,  using the proposed truncated rectified linear unit (TReLU) or Tanh as an activation function in the output layer of any deep neural network, see \eqref{eqn:margic_act} in Section \ref{sec:model_image}.

Next, we define an effective $r^2$-discriminative localizer $\bm \delta^0$ as the one minimizing $J(\bm \delta)$ among all $r^2$-discriminative localizers. Then $\bm \delta^0$ can be regarded as an optimal localizer for identifying discriminative features to \textit{interpret} a learner's predictability through \textit{effective} disruption. 

\begin{definition}[Effective $\bm{r}^2$-discriminative] 
\label{def:ideal_delta}
For $0 \leq r^2 \leq r^2_{\max}$, an effective $r^2$-discriminative localizer to $d$ is defined as
\begin{equation}
\label{effective}
\bm{\delta}^0 \in \argmin_{\bm{\delta} \in \mathcal{H}_b: R^2(d, \bm{\delta}) \geq r^2} \ J(\bm{\delta}),
\end{equation}
where $\mathcal{H}_b$ is a candidate collection of localizers such that $\sup_{\bm{x}} \| \bm{\delta}(\bm{x}) \|_\infty \leq 1$, and we say that the localized features by $\bm{\delta}^0(\cdot)$ is effective $r^2$-discriminative.
\end{definition}

As noted in Definition \ref{def:ideal_delta}, $\bm{\delta}^0$ is a most effective localizer that minimizes the regularization $J(\cdot)$ among all $r^2$-discriminative localizers. Without loss of generality, we assume that $\bm{\delta}^0$ always exists\footnote{Otherwise, the definition can be adapted to an $\varepsilon$-global minimizer, where the difference between its minimum value and the global minimum is no less than or equal to $\varepsilon$.} but may not be unique in the sequel. Note that in the presence of multiple global minimizers in \eqref{effective}, each of them could be useful, since our goal is to estimate such an effective $r^2$-discriminative localizer.


To identify an effective discriminative localizer for a learner $d$, we maximize $R^2(d, \bm \delta)$ or the prediction risk $\mathbb{E} \big( L\big( d(\bm{X} - \bm{\delta}(\bm{X}) ), Y\big) \big)$ with respect to $\bm \delta$, under the restriction of $J(\bm \delta)$. This leads to our proposed framework: 
\begin{equation}
\label{eqn:pop_model}
\max_{\bm{\delta} \in \mathcal{H}_b} \ \mathbb{E} \big( L\big( d(\bm{X} - \bm{\delta}(\bm{X}) ), Y\big) \big), \ \text{subject to} \ J(\bm{\delta}) \leq \tau,
\end{equation}
where $\tau >0$ is a tuning parameter to balance the objective of deteriorating the prediction performance and magnitude of a localizer $\bm{\delta}(\cdot)$. 
To make the constraint sensible, we let $\tau \leq p$ since $\sup_{\bm{\delta} \in \mathcal{H}_b} J(\bm{\delta}) = p$. As shown in Lemma \ref{lemma:fisher}, a most effective $r^2$-discriminative localizer $\bm \delta^0$ can be identified by \eqref{eqn:pop_model}.

\begin{lemma}
\label{lemma:fisher}
Let $\bm{\delta}^0_{\tau}$ be a global maximizer of \eqref{eqn:pop_model}, and 
$$\tau^0 = \min \{ \tau \in (0,p]: R^2(d, \bm{\delta}^0_{\tau}) \geq r^2 \},$$
then $\bm{\delta}^0_{\tau^0}$ is an effective $r^2$-discriminative localizer with $J(\bm{\delta}^0_{\tau^0}) = \tau^0$.
\end{lemma}

Lemma \ref{lemma:fisher} says that \eqref{eqn:pop_model} recovers an effective $r^2$-discriminative localizer defined in Definition \ref{def:ideal_delta} in a similar fashion as Fisher consistency in classification \citep{lin2004note,bartlett2006convexity}. 

Given a training sample $(\bm{x}_i, y_i)^n_{i=1}$, we propose an empirical risk function to estimate $\bm{\delta}^0_{\tau}$ and $\tau^0$:
\begin{align}
\label{eqn:emp_model}
 \max_{\bm{\delta} \in \mathcal{H}_b} \ L_n(d, \bm{\delta}) & = \frac{1}{n} \sum_{i=1}^n  L\big(d\big(\bm{x}_i - \bm{\delta}(\bm{x}_i) \big), y_i\big), \quad \text{subj to,} \quad J(\bm{\delta}) \leq \tau.
\end{align}
Denote $\widehat{\bm{\delta}}_\tau$ as a maximizer of \eqref{eqn:emp_model} for a given $\tau$. In view of Lemma \ref{lemma:fisher}, our final estimate of $(\bm{\delta}^0_{\tau^0}, \tau^0)$ is
\begin{align}
\label{eqn:sol}
& \widehat{\bm{\delta}}_{\widehat{\tau}} \text{ is a maximizer of \eqref{eqn:emp_model}}, \quad \text{where} \ \widehat{\tau} = \min \{ \tau \in (0,p]: R^2(d, \widehat{\bm{\delta}}_{\tau}) \geq r^2 \}.
\end{align}
In practice, $\tau \in (0,p]$ is replaced by $\tau \in \bm{\tau}$, where $\bm{\tau}$ is the candidate set of the tuning parameter $\tau$ as some grid points for positive real numbers, and the estimated $R^2$ is evaluated based on an independent test sample $\mathcal{D}_{\text{test}} = (\bm{x}_i, y_i)^{n+m}_{i=n+1}$,
\begin{equation}
  \label{eqn:est_r_sqare}
\widehat{R}^2( d, \widehat{\bm{\delta}}_\tau; \mathcal{D}_{\text{test}}) = 1 - \frac{ \sum_{(\bm{x}, y) \in \mathcal{D}_{\text{test}} }  L ( d(\bm{x}), y ) }{ \sum_{(\bm{x}, y) \in \mathcal{D}_{\text{test}}}  L ( d(\bm{x} - \widehat{\bm{\delta}}_\tau(\bm{x})), y ) }.
\end{equation}
Taken together, we iteratively solve \eqref{eqn:emp_model} for $\tau \in \bm{\tau}$ from the smallest to the largest via a grid search \citep{bergstra2012random}, and it terminates once $\widehat{R}^2( d, \widehat{\bm{\delta}}_\tau; \mathcal{D}_{\text{test}})$ exceeds a prespecified target $r^2$.

\subsection{A convolutional auto-encoder discriminative localizer}
\label{sec:model_image} The proposed framework \eqref{eqn:emp_model} admits a general localizer, such as a deep neural network. In practice, a network architecture incorporating data structure would be preferred \citep{bengio2012practical}. For example, for the image-to-image localization in the MNIST dataset, or the sequence-to-sequence localization in the ECG dataset, convolutional auto-encoder architectures are natural options to impose a ``local smoothing'' structure of the localized features. Therefore, this section illustrates the localizer $\bm{\delta}$ as a convolutional auto-encoder. It is noted that the network architecture of a discriminative model sets a standard for designing 
a localizer's architecture.

Consider a localizer of the form $\bm{\delta}(\bm{x}) =  \bm{x} \odot \bm{\pi}(\bm{x})$, $\bm{x}$ is an image, where $\odot$ is the element-wise product and $0 \leq \bm{\pi}(\bm{x}) \leq 1$ represents the percentage of image features that a localizer removes from the original feature $\bm{x}$. 

\begin{figure*}
  \centering
    \includegraphics[width=1.\textwidth]{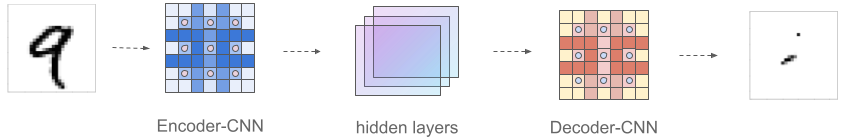}
   \caption{Our localization network structure based on a convolutional auto-encoder, which is composed of three components: Encoder-CNN (E-CNN), hidden neural network (HNN), and Decoder-CNN (D-CNN). }
   \label{fig:CAE}
\end{figure*}

Subsequently, we implement our proposed localizer by taking an image $\bm x$ as input and giving output as disruption proportion $\bm{\pi}(\bm{x})$. Specifically, we build a convolutional auto-encoder discriminative localizer based on a convolutional auto-encoder network (CAE; \cite{masci2011stacked, rumelhart1985learning}), which is composed of three components: Encoder-CNN (E-CNN), hidden neural network (HNN), and Decoder-CNN (D-CNN), as illustrated in Figure \ref{fig:CAE}. Besides, on the CAE backend model, we introduce a TReLU-softmax or Tanh+ReLU-softmax activation function to control the activity $L_1$-regularizer of the localizer. 
On this ground, we consider a localizer class:
\begin{align}
\label{eqn:funs_class}
\mathcal{H}^\tau_b = \Big\{ \bm{\delta}_\tau(\bm{x}) = \bm{x} \odot \bm{\pi}_{\bm{\theta}}^\tau(\bm{x}): \bm{\pi}_{\bm{\theta}}^\tau(\bm{x}) = A_\tau \big( \text{CAE}_{\bm{\theta}}(\bm{x}) \big); \ \bm{\theta} \in \bm{\Theta} \Big\},
\end{align}
where $\text{CAE}_{\bm{\theta}}(\bm{x})$ is a convolutional auto-encoder with
$\bm{\theta} \in \mathbb{R}^q$ denoting its parameters,
$\bm{\Theta}$ is a parameter space of $\bm{\theta}$, and $A_\tau(\cdot)$ is a structured activation function, such as:
\begin{align}
  \label{eqn:margic_act}
  A_\tau(\bm{z}) = \text{TReLu} \big( \tau \cdot \text{softmax}(\bm{z}) \big), \text{ or } A_\tau(\bm{z}) = \text{Tanh} \Big( \text{ReLu} \big( \tau \cdot \text{softmax}(\bm{z}) \big) \Big)
\end{align}
where $\text{TReLu}(\bm{u}) = \min(\bm{u}_+, \bm{1})$ is the truncated ReLU function.


Note that for any $\bm{\delta} \in \mathcal{H}^\tau_b$, based on the definition of $A_\tau$, the following conditions are automatically satisfied: (i) $\sup_{\bm x} \| \bm{\delta}(\bm{x}) \|_\infty \leq 1$; (ii) $J(\bm{\delta}) = \sup_{\bm{x}} \| \bm{\delta}(\bm{x}) \|_1 \leq \tau$. 
Therefore, the constraints in \eqref{eqn:emp_model} can be removed given $ \mathcal{H}^\tau_b$, and the optimization of \eqref{eqn:emp_model} becomes:
\begin{align}
\label{eqn:param_model}
& \max_{\bm{\theta}} \ \frac{1}{n} \sum_{i=1}^n  L\big(d\big(\bm{x}_i - \bm{x}_i \odot \bm{\pi}^\tau_{\bm{\theta}}(\bm{x}_i) \big), y_i\big), 
\end{align}
which can be solved by Gradient Descent (GD) or stochastic gradient descent (SGD; \cite{raginsky2017non}).
The GD solution of \eqref{eqn:param_model} attains a local maximizer of \eqref{eqn:param_model} under some mild assumptions \citep{lee2016gradient}. Note that the convergence result can be extended to SGD as in  \citep{ge2015escaping}, and a global maximizer may be obtained by GD or SGD with additional assumptions \citep{raginsky2017non}. Once $\widehat{\bm{\theta}}$ is obtained, the estimated localizer is specified as 
\begin{equation}
\label{eqn:estimator}
\widehat{\bm{\delta}}_\tau(\bm{x}) = \bm{x} \odot A_\tau \big( \text{CAE}_{\widehat{\bm{\theta}}}(\bm{x}) \big).
\end{equation}


\subsection{Interpretation uncertainty}

Robustness is a general challenge to existing interpretation approaches. For example, \cite{ghorbani2019interpretation} indicates that systematic perturbations can lead to dramatically different interpretations without changing the label. 
To distinguish the interpretability and robustness for the proposed framework, we propose an \textit{unexplainable $R^2$} as a confidence interval for the generalized partial $R^2$ to distinguish the prediction deterioration caused by discriminative features from model instability. In particular, given a learner $d$ and a localizer $\widehat{\bm{\delta}}_\tau$, we construct a confidence interval for $R^2( d, \widehat{\bm{\delta}}_\tau)$ via bootstrap on a test sample. 

\begin{figure}[ht]
  \centering
    \includegraphics[width=.6\textwidth]{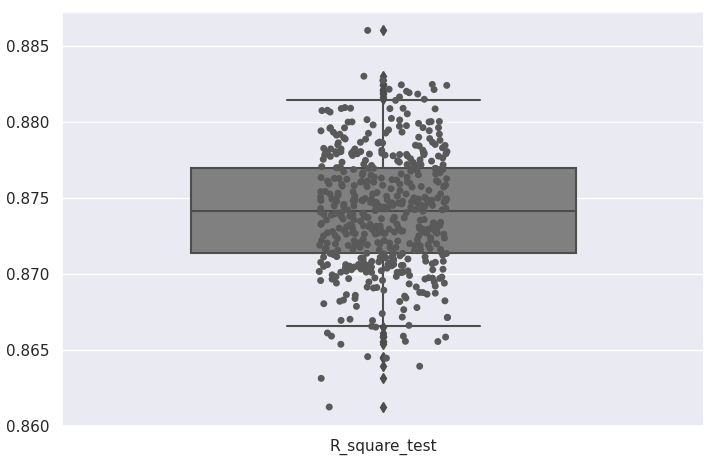}
    \caption{Boxplot of the estimated $R^2$ of the proposed method based on 500 bootstrap samples for the MNIST benchmark example. This example illustrates the concept of an unexplainable $R^2$.}
   \label{fig:boxplot}
\end{figure}

First we generate a bootstrap sample $\mathcal{D}^{(b)}_{\text{test}}$ by drawing $B$ independent observations from the test data $\mathcal{D}_{\text{test}}$ with replacement.
Then the unexplainable $R^2$ for $R^2( d, \widehat{\bm{\delta}}_\tau)$ is obtained using the sampling distribution of the bootstrapped estimates $\big \{\widehat{R}^2( d, \widehat{\bm{\delta}}_\tau; \mathcal{D}^{(B)}_{\text{test}}) \big \}_{b=1}^B$. For example, for the MNIST dataset, we obtain a 95\% confidence interval of $R^2( d, \widehat{\bm{\delta}}_\tau)$ by computing the $\floor{.025B}$-th and $\floor{.975B}$-th ordered estimated $R^2$ on the bootstrap samples, as indicated in Figure \ref{fig:boxplot}. More detail can be found in Section \ref{sec:application}.


\section{Theoretical guarantee}

This section indicates that the proposed framework yields discriminative features attaining a target $R^2$ with optimal effectiveness asymptotically.
\label{sec:rate}

To proceed, let $\bm{\delta}^0_{\tau}$ be a global maximizer of \eqref{eqn:pop_model} over a function class
$
\mathcal{H}_b = \big\{ \bm{\delta} \in \mathcal{H}: \ \sup_{\bm{x}} \| \bm{\delta}(\bm{x}) \|_\infty \leq 1 \big\}.
$
Without loss of generality, assume that $0 \leq L(d(\bm{x}_{\bm{\delta}}), Y) \leq U$ for a sufficiently large constant $U \geq 1$, for any $\bm{\delta} \in \mathcal{H}_b$ and $\bm{x} \in \mathbb{R}^p$ \citep{wu2007robust}. 
To make the constraint sensible, we let $\tau \leq p$ since $\sup_{\bm{\delta} \in \mathcal{H}_b} J(\bm{\delta}) = p$.

Denote the Rademacher complexity for the function class $\mathcal{H}_{b}$ as $\kappa_{n} = \mathbb{E} \mathcal{R}_n\big( \mathcal{H}_{b} \big) = \sup_{\bm{\delta} \in \mathcal{H}_{b}} n^{-1} \sum_{i=1}^n \big| \eta_i \big( L(d(\bm{X}_i - \bm{\delta}(\bm{X}_i)), Y) \big) \big|$, and $\{\eta_i\}_{i=1}^n$ are i.i.d. Rademacher random variables with $\eta_i$ taking the values +1 and -1 with probability 1/2 each. To make the constraint sensible, we let $\tau \leq p$ since $\sup_{\bm{\delta} \in \mathcal{H}_b} J(\bm{\delta}) = p$.
Theorem \ref{thm:sup_asymp_r2} gives a convergence rate for the discrepancy between $\bm{\delta}^0_\tau$ and $\widehat{\bm{\delta}}_\tau$ in terms of $R^2$ uniformly over $0 < \tau \leq p$.

\begin{theorem}[Asymptotics of $\bm{R}^2$]
\label{thm:sup_asymp_r2}
Let $\widehat{\bm{\delta}}_\tau$ be a global maximizer of \eqref{eqn:emp_model}, for $\varepsilon_n \geq 8\kappa_{n}$ and
any predictive model $d$, we have
\begin{equation*}
\mathbb{P} \big( \sup_{ 0 < \tau \leq p} R^2(d, \bm{\delta}^0_\tau) - R^2(d, \widehat{\bm{\delta}}_\tau) \geq \varepsilon_n \big) \leq K \exp \big( - \frac{n \varepsilon_n^2}{KU^2} \big),
\end{equation*}
where $K>0$ is a constant. Hence, $$ \sup_{0 < \tau \leq p} (R^2(d, \bm{\delta}^0_\tau) - R^2(d, \widehat{\bm{\delta}}_\tau)) = O_p \big( \max( \kappa_{n}, n^{-1/2} ) \big).$$ 
\end{theorem}

Note that the asymptotics of the Rademacher complexity $\kappa_n$ for a candidate class $\mathcal{H}$ has been extensively investigated in the literature \citep{bartlett2002rademacher, bartlett2005local}. Therefore, the uniform convergence rate can be obtained for a generic candidate class by Theorem \ref{thm:sup_asymp_r2}. 
Moreover, the asymptotics for a fixed $\tau$ is also provided in Appendix C, where the rate can be further improved. 

Next, we show that $\widehat{\bm{\delta}}_{\widehat{\tau}}$ is an asymptotically effective $r^2$-discriminative localizer. Note that $\widehat{\bm{\delta}}_{\widehat{\tau}}$ already is an $r^2$-discriminative localizer, since $R^2(d, \widehat{\bm{\delta}}_{\widehat{\tau}}) \geq r^2$ by the definition of $\widehat{\tau}$ in \eqref{eqn:sol}. Therefore, it suffices to show \textit{effectiveness}, that is,
that is, $|J(\widehat{\bm{\delta}}_{\widehat{\tau}}) - J(\bm{\delta}^0_{\tau^0})| = | \widehat{\tau} - \tau^0 | \overset{p}{\longrightarrow} 0$. 
To proceed, we require a smoothness condition of $R^2(d, \bm{\delta}^0_\tau)$ over $\tau$ in Assumption A.

\noindent \textbf{Assumption A} (Smooth). Assume that $R^2(d, \bm{\delta}^0_\tau)$ is a continuous function in
$\tau$. Moreover, there exists a constant $\mu_0 > 0$ such that $|\tau_1 - \tau_2| \leq \mu$ if 
$| R^2(d, \bm{\delta}^0_{\tau_1}) - R^2(d, \bm{\delta}^0_{\tau_2})| \leq c_0 \mu^\alpha$ for any 
$\mu \leq \mu_0$.

\begin{theorem}[Oracle property] 
\label{thm:oracle}
Let $\bm{\delta}^0$ be an effective $r^2$-discriminative localizer in Definition \ref{def:ideal_delta} and $\widehat{\bm{\delta}}_{\widehat{\tau}}$ be
a global maximizer of \eqref{eqn:sol}. Under Assumption A, for $\omega_n \geq 2(8\kappa_n / c_0)^{1/\alpha}$, we have
\begin{align*}
  \mathbb{P} \Big( \big | \widehat{\tau} - \tau^0 \big| \geq \omega_n \Big)  & = \mathbb{P} \Big( \big |J(\widehat{\bm{\delta}}_{\widehat{\tau}}) - J( \bm{\delta}^0 ) \big| \geq \omega_n \Big) \leq K' \exp\Big( - \frac{n \omega_n^{2\alpha}}{K' U^2} \Big),
\end{align*}
where $K' > 0$ is a universal constant.
Therefore,
$$
|J(\widehat{\bm{\delta}}_{\widehat{\tau}}) - J(\bm{\delta}^0_{\tau^0})| = | \widehat{\tau} - \tau^0 | \overset{p}{\longrightarrow} 0,
$$
and $\widehat{\bm{\delta}}_{\widehat{\tau}}$ is an asymptotically effective $r^2$-discriminative localizer. 
\end{theorem}

Therefore, the proposed framework yields an effective $r^2$-discriminative localizer as defined in \eqref{def:ideal_delta}, rendering theoretical reliable discriminative features for a target $R^2$. 
Moreover, the theorems are illustrated for the proposed convolutional auto-encoder neural network \eqref{eqn:param_model} in Corollary B.1, where the convergence rates are computed depending on the sample size and the network architecture.

\section{MNIST benchmark}
\label{sec:application}

This section examines the numerical performance and visualizes discriminative features generated from the proposed localizer for the MNIST handwritten digit dataset \citep{lecun-mnisthandwrittendigit-2010} (\url{http://yann.lecun.com/exdb/mnist/}).  
All empirical results are produced in our Python library \pkg{dnn-locate} (\url{https://github.com/statmlben/dnn-locate}).

For the MNIST data, we extract 14,251 images (28$\times$28 field) from the dataset with labels `7' and `9'.  Our goal is to localize discriminative features for distinguishing digits `7' and `9' with a specific generalized partial $R^2$.

First, we train a decision function $d$ as a CNN, where we regularize each parameter of the CNN by the $L_1$-norm with weight 0.001.
Here the CNN model is optimized by the Adam algorithm with an initial learning rate of 0.001, early stopping based on the validation accuracy with patience as 10, and 20\% of the training data as a validation set. 


Then, a convolutional auto-encoder (CAE), as in \eqref{eqn:funs_class} and Figure \ref{fig:CAE}, is constructed 
as the localizer. 
For training, we optimize the model by stochastic gradient descent with an initial learning rate of $10/\tau$ and reduce the learning rate by a factor of 0.382 \citep{bengio2012practical}, when the validation loss has stopped improving. Moreover, early stopping is conducted based on validation accuracy with patience as 15 \citep{raskutti2014early}. 

\begin{figure}
  \centering
    \includegraphics[width=0.85\textwidth]{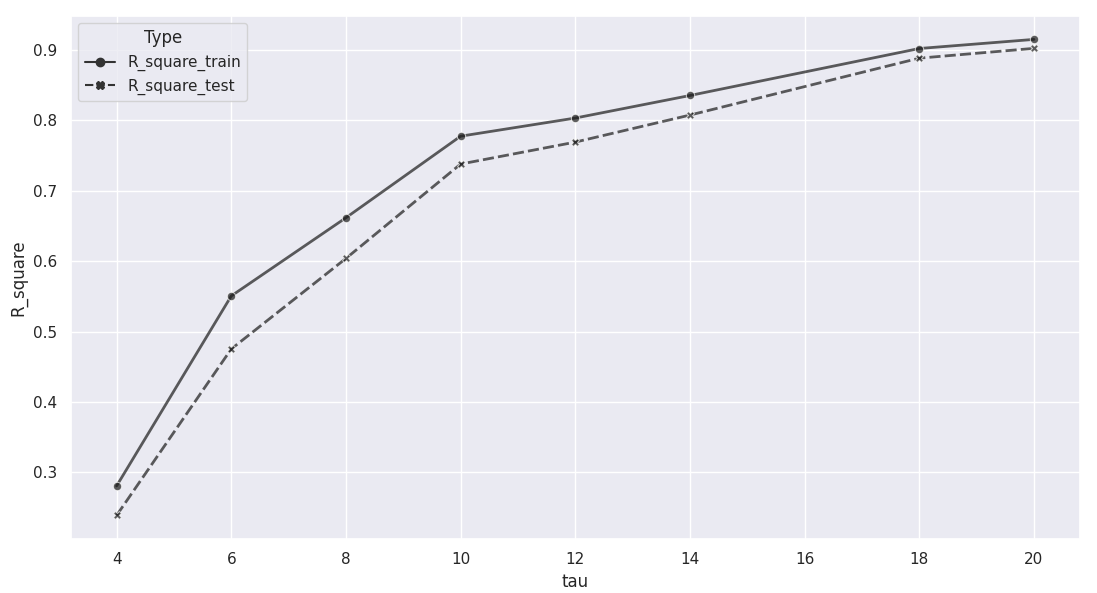}
   \caption{Training and testing estimated $R^2$s for the proposed framework in handwritten digit dataset with $\tau = 6, 8, 10, 12, 14, 16, 18, 20$, which indicates that the $R^2$ increases as the magnitude for an estimated localizer becoming large. }
   \label{fig:tau_r2}
\end{figure}
\begin{figure*}
  \centering
    \includegraphics[width=.9\textwidth]{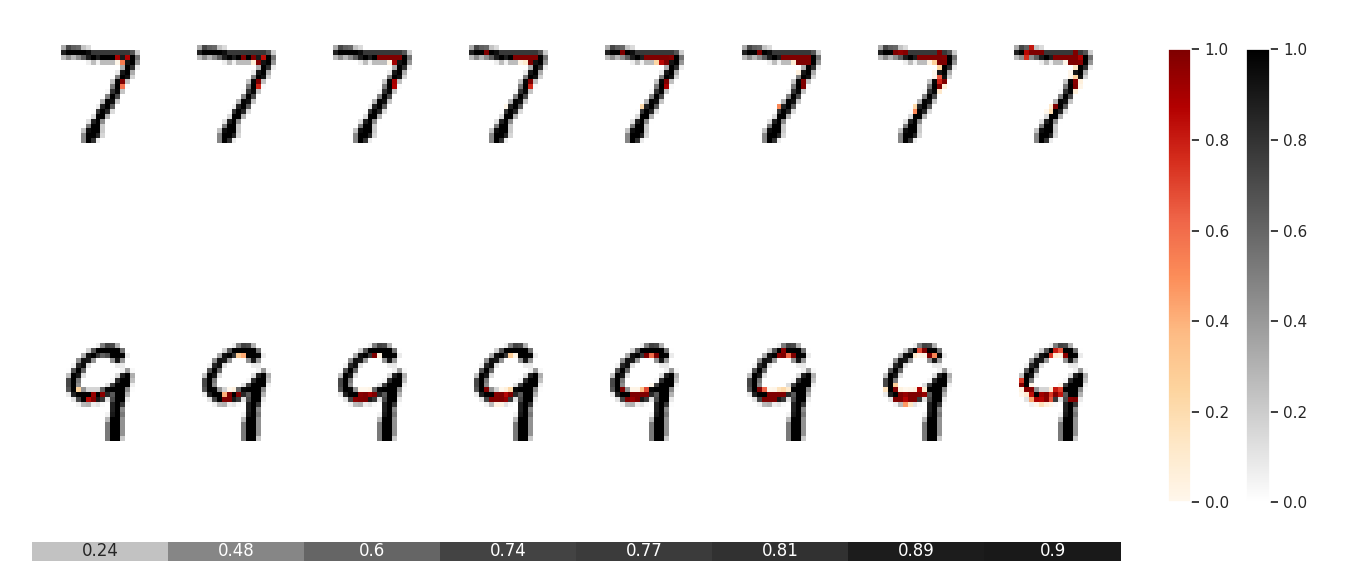}
   \caption{Illustrative instances of localized discriminative features (red) by the proposed method for `7' and `9' digits (black) as well as their corresponding estimated $R^2$s (the heatmap in x-axis). The gray color-bar indicates gray scale of original images, and the red color-bar indicates the proportion of removing features, that is, $\bm{\pi}(\bm{x})$ in \eqref{eqn:param_model}.}
   \label{fig:demo}
\end{figure*}

For the proposed method, we implement \eqref{eqn:param_model} based on $\tau = 4, 6, 8, 10, 12, 14, 18, 20$, and the relation between $\tau$ and its corresponding estimated $R^2$s are demonstrated in Figure \ref{fig:tau_r2}. Note that the estimated $R^2$ increases as the activity $L_1$-norm of the localizer becoming large.
Furthermore, the discriminative features, identified by the proposed
method for two illustrative instances of `7' and `9', are visualized in Figure \ref{fig:demo}.
Specifically, as the estimated $R^2$ becomes larger, the disrupted instance labeled as `9' becomes more and more like `7'. 

As illustrated by the boxplot (Figure \ref{fig:boxplot}), a 95\% confidence interval [0.867, 0.882] for the $R^2(d, \widehat{\bm{\delta}}_\tau)$ indicates some uncertainty with the fitted localizer ($\tau = 17$), where the $R^2$ is categorized as unexplainable if it falls inside the confidence interval. 
 
Next, we compare the proposed method with five state-of-the-art methods by both human visual and numerical evaluations, including deep Taylor explainer \citep{montavon2017explaining}, gradient-based explainer \citep{selvaraju2017grad}, lrp.z \citep{bach2015pixel}, deconvnet \citep{zeiler2014visualizing}, and pattern.net \citep{kindermans2017learning}. 
All competitors are implemented by the Python library \texttt{innvestigate} (\url{https://github.com/albermax/innvestigate}), and the \texttt{batch size} is set as 64 for pattern.net.
In particular, a heatmap of discriminative features produced by each method is validated by a visual inspection and by a numerical comparison based on the estimated $R^2$ given the same amount/magnitude of feature disruption. 

\subsection{Visual comparison}

\begin{figure*}
  \centering
    \includegraphics[width=.85\textwidth]{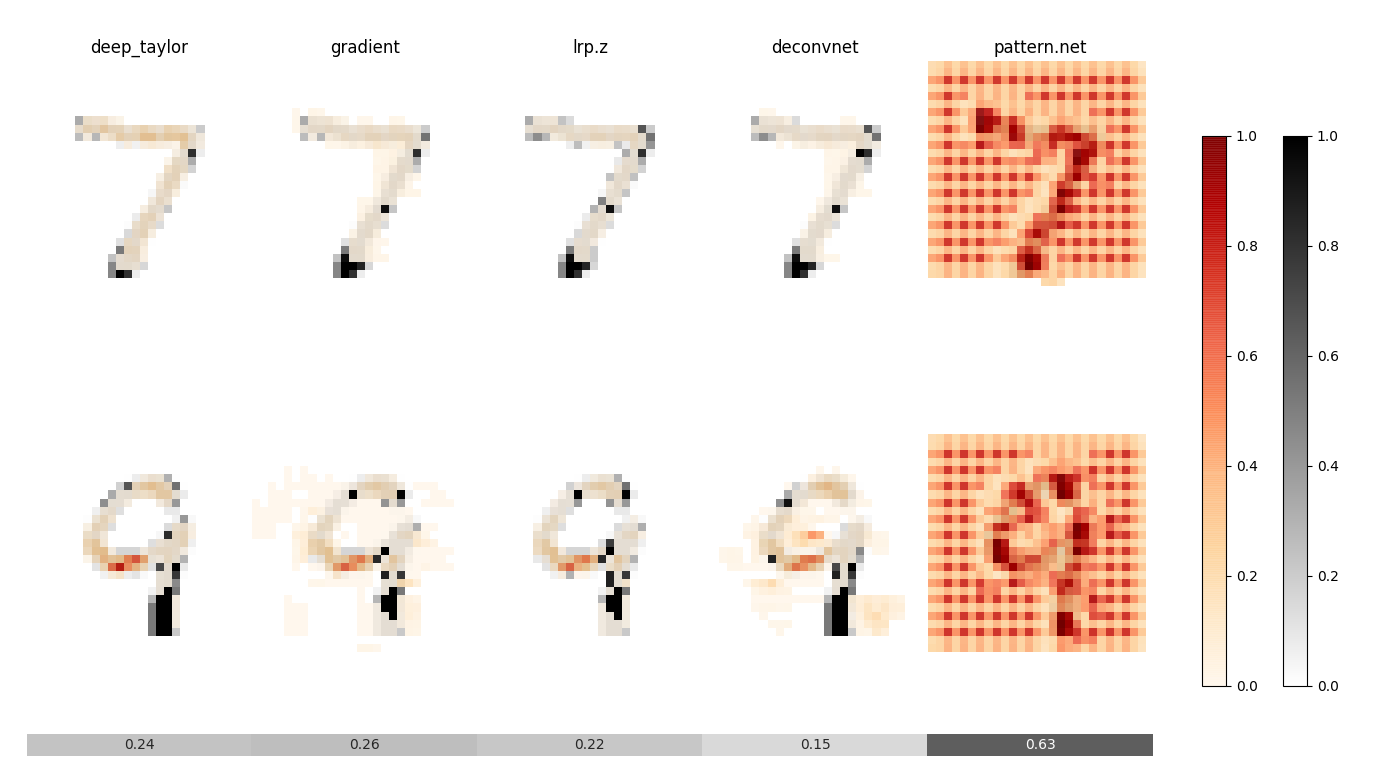}
   \caption{Illustrative instances of localized discriminative features (red), based on five competitors, for `7' and `9' digits (black), and their corresponding generalized partial $R^2$s (the heatmap in x-axis). The gray color-bar indicates gray scale of original images, and the red color-bar indicates importance of pixels produced by a localizer.}
   \label{fig:camp_competitor}
\end{figure*}

As displayed in Figure \ref{fig:demo}, the proposed method produces more compact discriminative features. By comparison, the other competitors yield dense image features spreading over the entire digits. Moreover, the proposed method gives roughly equal attention to two images in discriminating digits '7' from '9', which conforms with human intuition. However, as depicted in \ref{fig:camp_competitor}, all competitors generate imbalanced discriminative features that are more in one of the two images of `7' and `9' as shown in Figure \ref{fig:camp_competitor}.
As a result, the proposed method is more conducive for label-specific analysis.

\subsection{Numerical comparison}

To make a fair comparison, we conduct a pairwise comparison between the proposed localizer and each competitor under the same magnitude of $J(\cdot)$. 
Specifically, we compute the value of $J(\cdot)$ and the estimated $R^2$ of detected regions by a competitor. To fairness, we chose our tuning parameter $\tau$ to be the same as the $J(\cdot)$ of the competitor. Then compare the $R^2$s for the proposed method and the corresponding competitor.

  As indicated in Table \ref{tab:perf}, under the same magnitude $J(\cdot)$, the proposed localizer 
outperforms all competitors in terms of $R^2$, where the amounts of improvement 
are 58.47\%, 147.1\%, 146.5\%, 308.0\%, and 44.14\%.

\begin{table*}[!h]\centering
  \caption{
    A pairwise comparison for the proposed framework and five existing methods based on 10-fold cross validation. Here $J(\cdot)$ is the activity $L_1$-regularizer as defined in \eqref{eqn:l1-norm}, and the estimated $R^2$ as in \eqref{eqn:est_r_sqare}.}
  \scalebox{.9}{
  \begin{tabular}{@{}cccccccccccccccccc@{}} \toprule
  \phantom{a} & ~ & \phantom{a} & activity $L_1$-norm $J(\cdot)$ & \phantom{a} & $\widehat{R}^2$ (competitor in the first column) & \phantom{a} & $\widehat{R}^2$ (our method) & \phantom{a} \\
  \midrule
   & deep-Taylor && 11.698(.228) && 0.236(.016) && 0.374(.084) \\
   & gradient && 26.028(.319) && 0.289(.012) && 0.714(.033) \\
   & lrp.z && 14.689(.219) && 0.256(.014) && 0.631(.077) \\ 
   & deconvnet && 27.832(.955) && 0.175(.015) && 0.714(.023) \\ 
   & pattern.net && 374.709(2.762) && 0.648(.006) && 0.934(.001) \\ 
  \bottomrule
  \end{tabular}}
  \label{tab:perf}
\end{table*}

In summary, the proposed method has significant benefits. First, as illustrated in Figure \ref{fig:demo}, it provides a flexible framework to localize desirable discriminative features to explain a certain amount of predictive performance as measured by an $R^2$. Second, the visual and numerical results in Figures \ref{fig:demo} and \ref{fig:camp_competitor} and Table \ref{tab:perf} suggest that the proposed method can produce compact and effective discriminative features, which are consistent with human visual judgment.

\section{ECG data analysis}
\label{sec:ECG}
Finally, we present the results of applying our method to the MIT-BIH Arrhythmia Electrocardiogram (ECG) dataset for heartbeat classification \citep{moody1990bih}.
The MIT-BIH dataset consists of ECG recordings from 47 different subjects recorded at the sampling rate of 360Hz by the BIH Arrhythmia Laboratory. Each beat is annotated into 5 different classes by following the Association for the Advancement of Medical Instrumentation (AAMI) EC57 standard: labeled as 'N', 'S', 'V', 'F', and 'Q'.
The pre-processed dataset is publicly available at \url{https://www.kaggle.com/shayanfazeli/heartbeat}. 
The MIT-BIH ECG dataset has been extensively studied, 
including using deep convolutional neural networks  \citep{kachuee2018ecg, acharya2017deep,martis2013application}. In spite of the impressive predictive performance obtained by the devised networks (with more than 93\% classification accuracy), it is unknown why and how the networks achieved their good performance. To advance our understanding and possibly offering new insights, our goal is to localize discriminative signal fragments based on the deep CNN developed in \cite{kachuee2018ecg}, which is one of the state-of-the-art ECG classification methods.

For implementation, we build a localizer by using a convolutional auto-encoder structure in Figure \ref{fig:CAE} with two convolutional layers as an encoder and two transposed convolution layers as a decoder. For training, we use the SGDW optimizer with ``\texttt{learning\_rate=.1}'', ``\texttt{weight\_decay=1e-4}'', ``\texttt{momentum=.9}''. Besides, a reducing learning rate scheme is used with ``\texttt{factor=.382}'' and ``\texttt{patience=3}'', and early stopping is adopted with ``\texttt{patience=20}''. Moreover, we tune the hyperparameter $\tau$ to achieve various $R^2$s: 10\%, 50\%, 60\%, 70\%, 75\%. Training one network takes less than half an hour on a GeForce GTX 2060Ti GPU. All the Python codes are publicly available in \url{https://github.com/statmlben/dnn-locate}.

To demonstrate our localization results, we concentrate on the localized ECG signals under the label `S' (including atrial premature, aberrant atrial premature, nodal premature, and supra-ventricular premature) and the label `V' (including premature ventricular contraction, and ventricular escape). 

\begin{figure}
  \centering
  \includegraphics[scale=0.15]{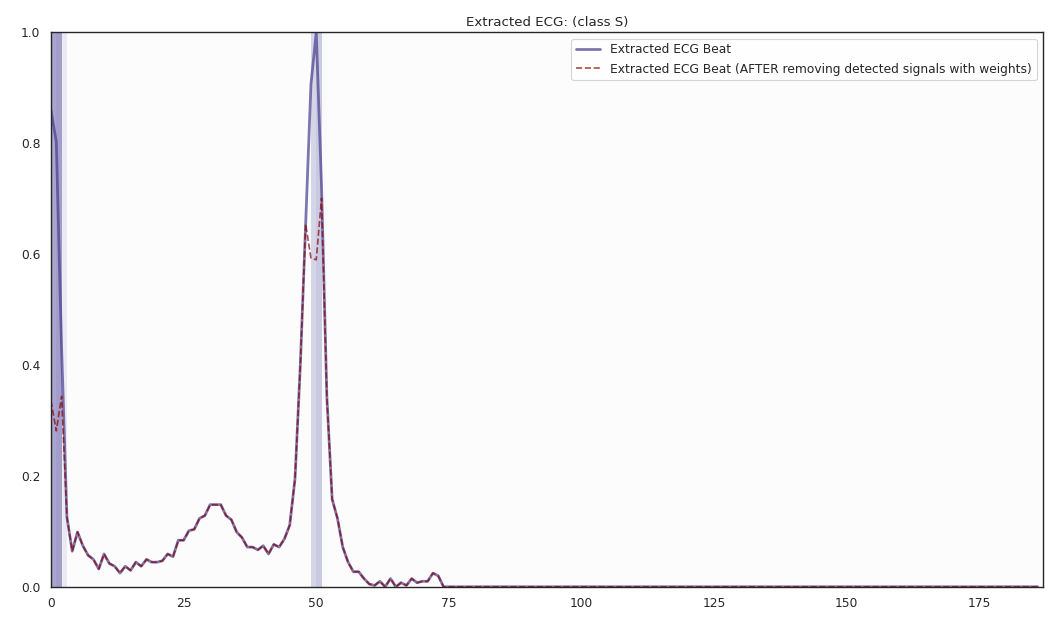}
  \includegraphics[scale=0.15]{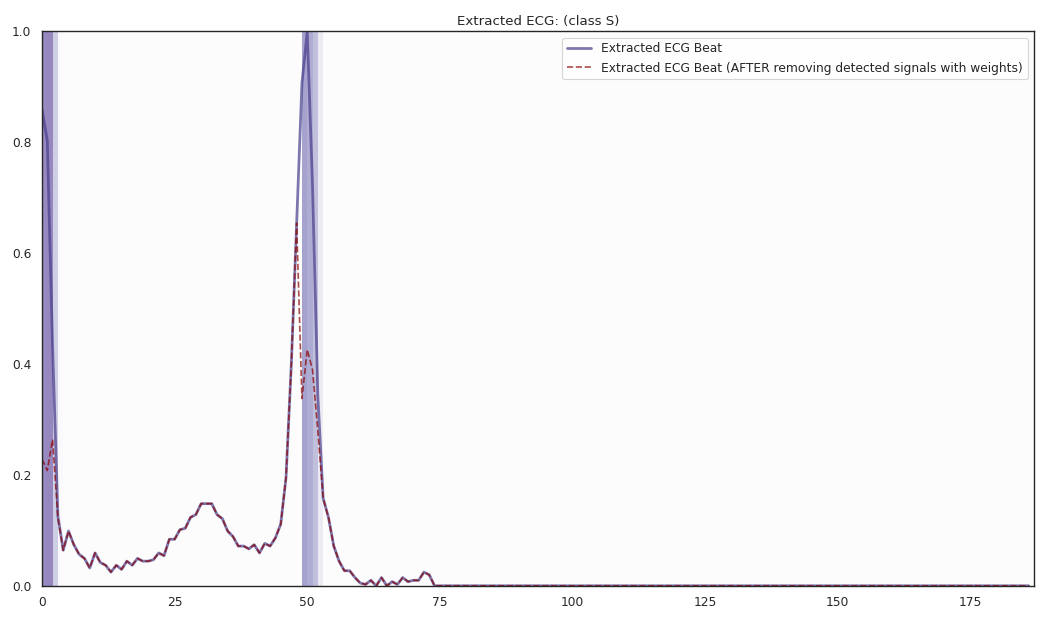}
  \includegraphics[scale=0.15]{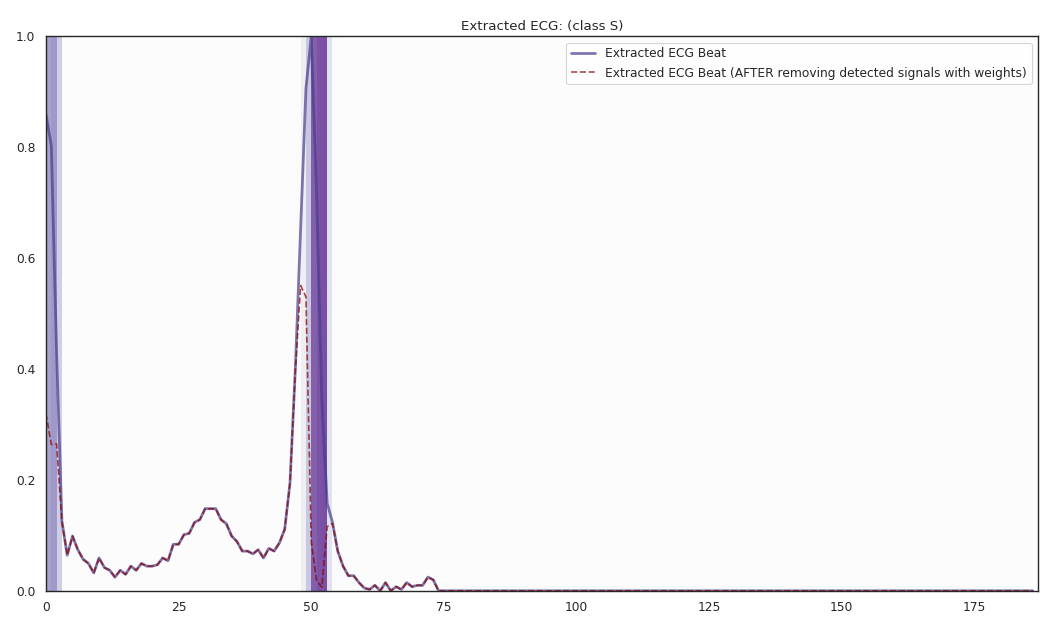}
  \includegraphics[scale=0.15]{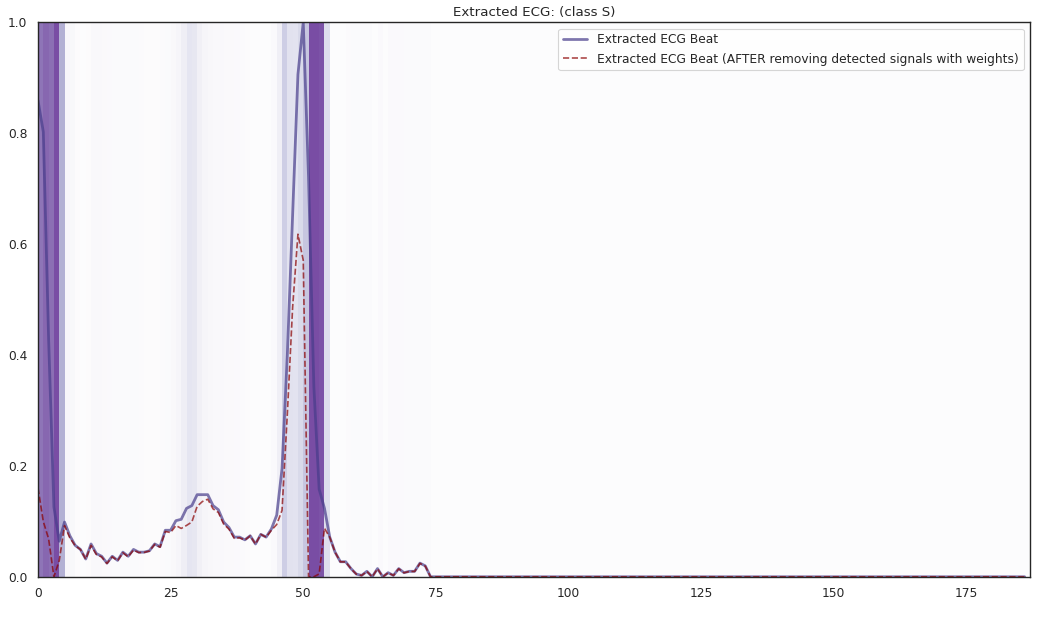}
  \includegraphics[scale=0.15]{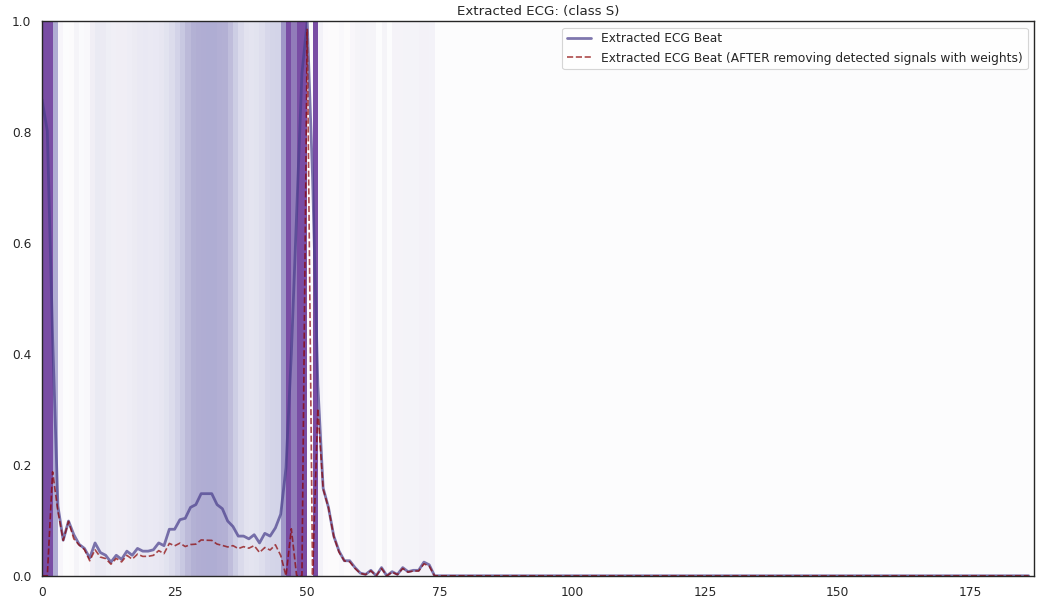}
  \includegraphics[scale=0.15]{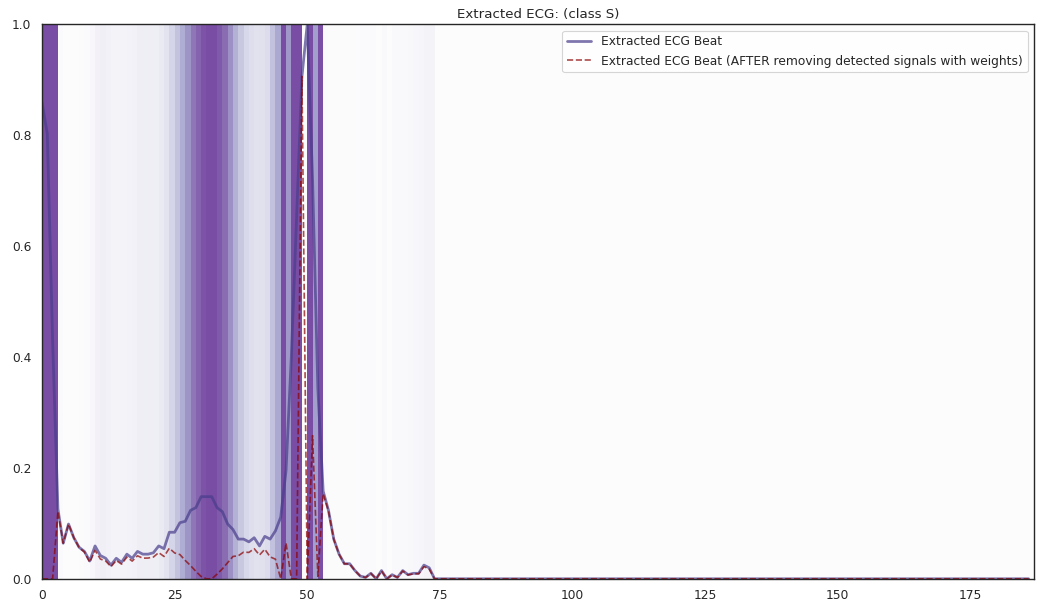}
  \rule[1ex]{\textwidth}{0.1pt}
  \includegraphics[scale=0.15]{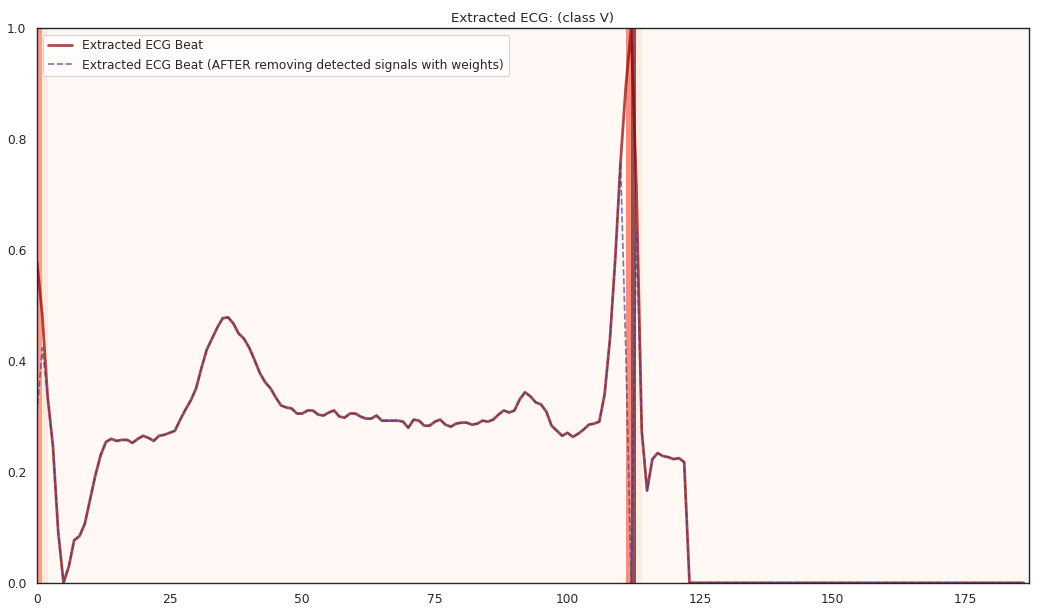}
  \includegraphics[scale=0.15]{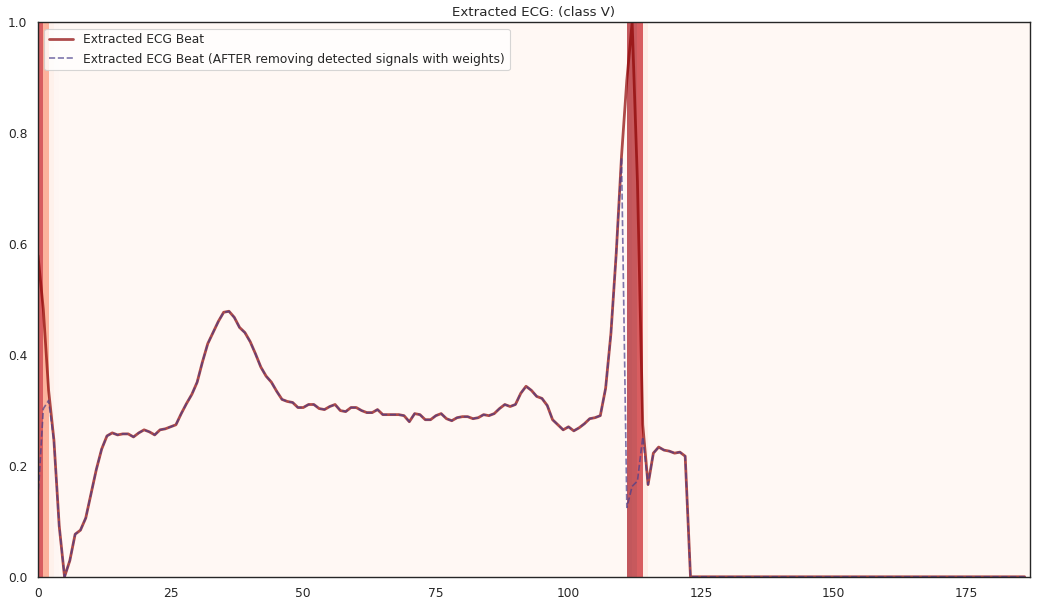}
  \includegraphics[scale=0.15]{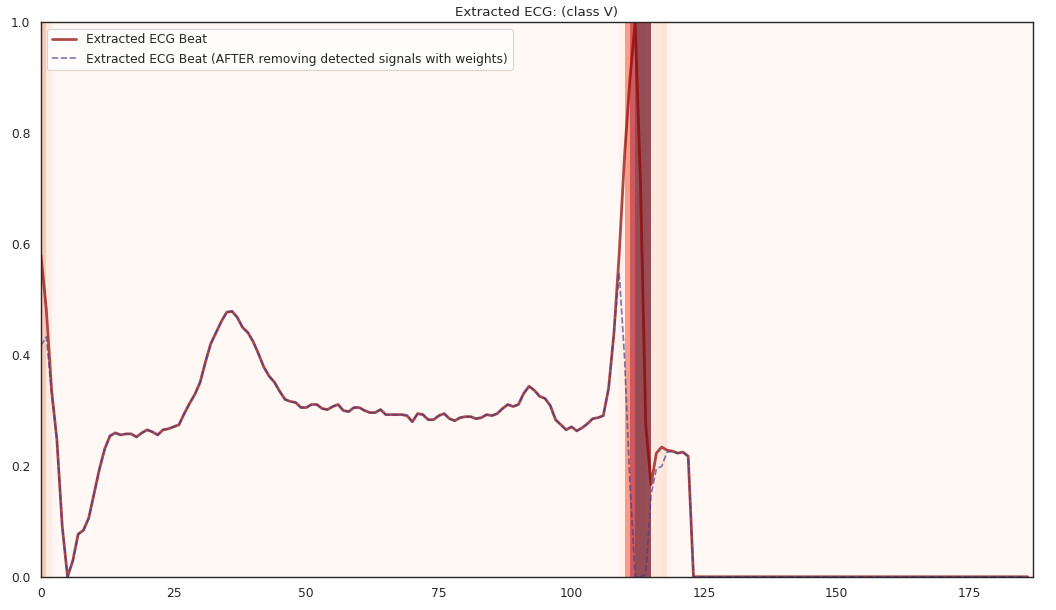}
  \includegraphics[scale=0.15]{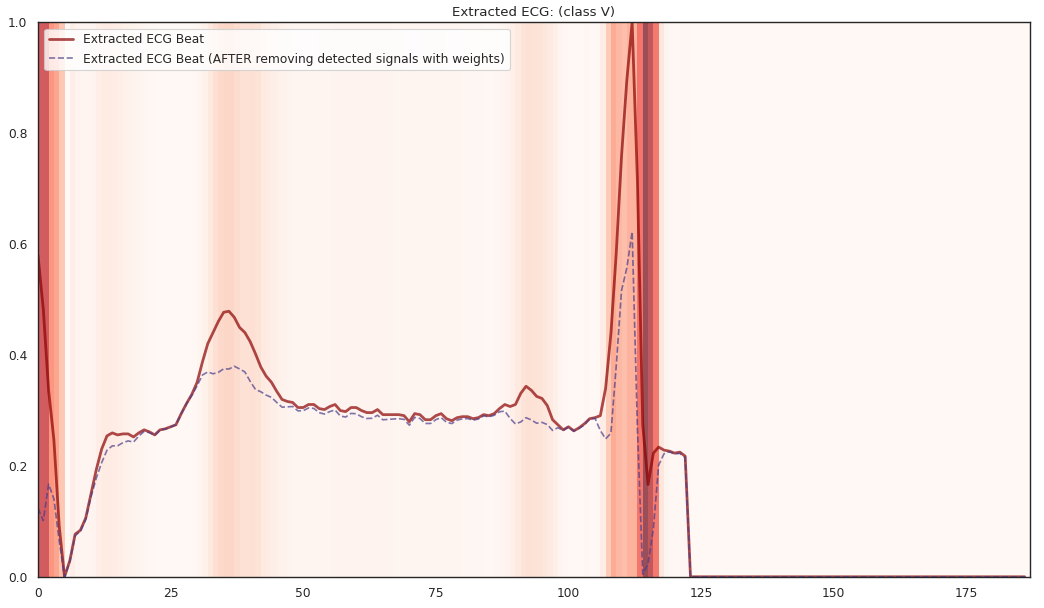}
  \includegraphics[scale=0.15]{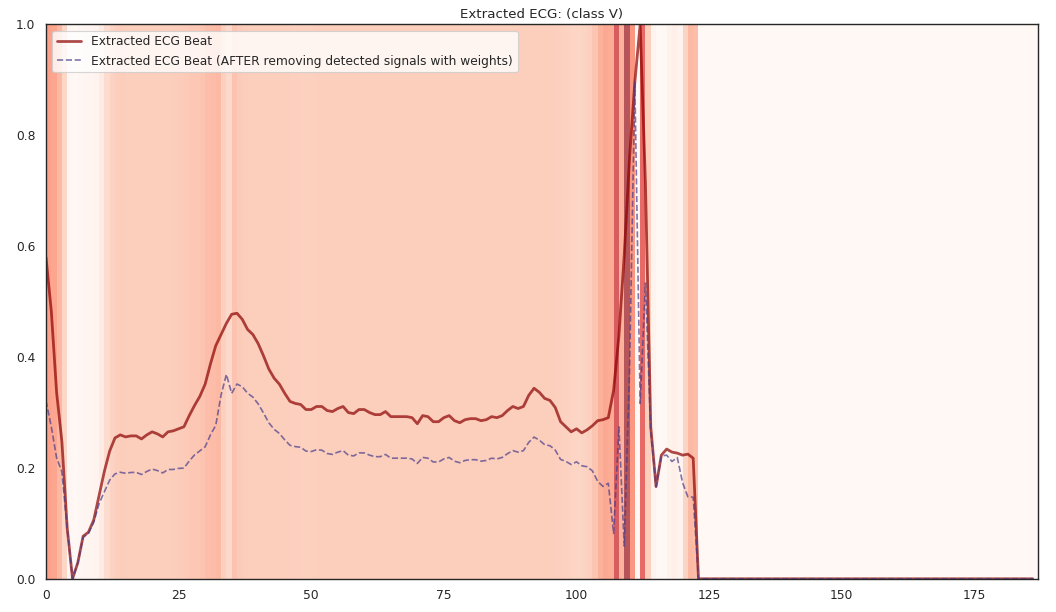}
  \includegraphics[scale=0.15]{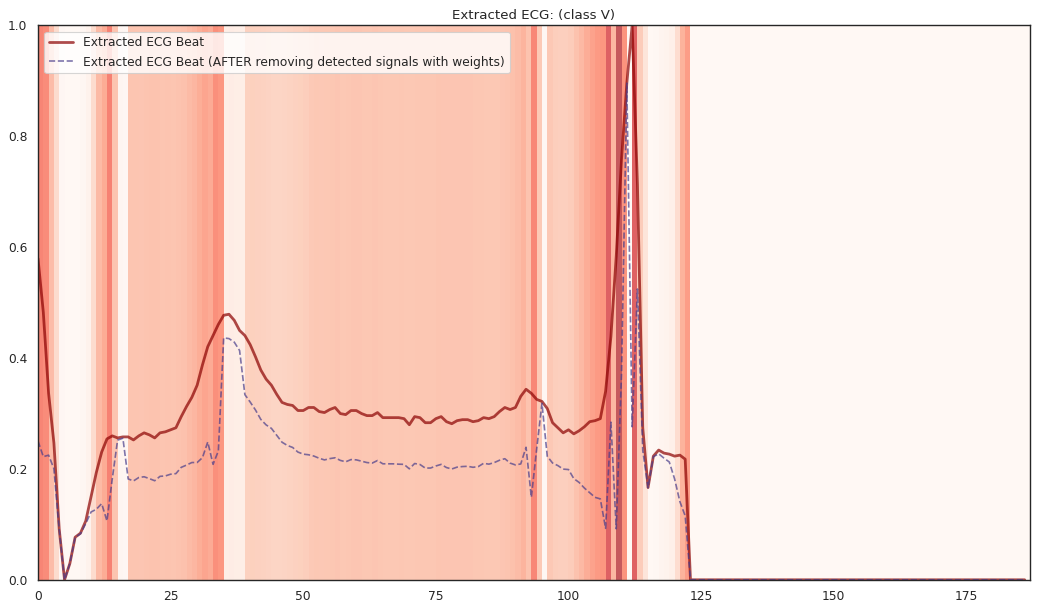}
  \caption{The proposed method reveals subtle discriminative features used by a deep convolutional neural network for ECG classification on the MIT-BIH dataset. The left/right panel highlights the localized features for a particular individual under label `V'/`S'. Note that the $R^2$ increases from 10\% to 88\% for both upper and lower panels. The medical literature provided in Section \ref{sec:ECG} gives supporting evidence for biological plausibility of the localized features; see more discussion in Section \ref{sec:ECG}.}
  \label{fig:ECG}
\end{figure}

As shown in the lower panel of Figure \ref{fig:ECG}, the localized regions (highlighted by the red bars) of ECG complexes in sinus rhythm are most informative in distinguishing presence of ventricular ectopic beats from supraventricular ectopic beats in a particular individual. 
The localized regions lie in the QRS complex, which correlates with ventricular depolarization or electrical propagation in the ventricles \citep{mirvis2001}. 
Ion channel aberrations and structural abnormalities in the ventricles can affect electrical conduction in the ventricles \citep{rudy2004ionic}, manifesting with subtle anomalies in the QRS complex in sinus rhythm that may not be discernible by the naked eye but is detectable by the convolutional auto-encoder. 
Of note, as the $R^2$ increases from 10\% to 88\%, the highlighted color bar is progressively broader, covering a higher
proportion of the QRS complex. 
The foregoing observations are sensible: the regions of interest resided in the QRS complex are biologically plausible and consistent with cardiac electrophysiological principles.

As shown in the upper panel of Figure \ref{fig:ECG}, similarly, the regions of interest (highlighted by the blue bars) of ECG complexes in sinus rhythm are most informative in distinguishing the presence of supraventricular ectopic beats from ventricular ectopic beats in a particular individual. As in the left panel, the regions of interest lies in the QRS complex, which is intuitive and biologically plausible as explained above.

As shown in the last three figures in the upper panel of Figure \ref{fig:ECG} for supraventricular complexes, as the $R^2$ increases from 80\% to 84\% and finally 88\%, the blue bar progressively highlights the P wave of ECG complexes in sinus rhythm. This observation is consistent with our understanding of the mechanistic underpinnings of atrial depolarization, which correlates with the P wave. Ion channel alterations and structural changes in the atria can affect electrical conduction in the atria \citep{rudy2004ionic}, manifesting with subtle anomalies in the P wave in sinus rhythm that may not be discernible by the naked eye but are detectable by the convolutional auto-encoder.

Collectively, the examples above underscore the fact that the discriminative regions of interest identified by our proposed method are biologically plausible and consistent with cardiac electrophysiological principles while locating subtle anomalies in the P wave and QRS complex that may not be discernible by the naked eye. 
By inspecting our results with an ECG clinician (Dr. Chen in the authorship), the localized discriminative features of the ECG are consistent with medical interpretation in ECG diagnosis.

\subsection{Robustness against localization network architecture} 
\begin{figure}
  \centering
  \includegraphics[scale=0.16]{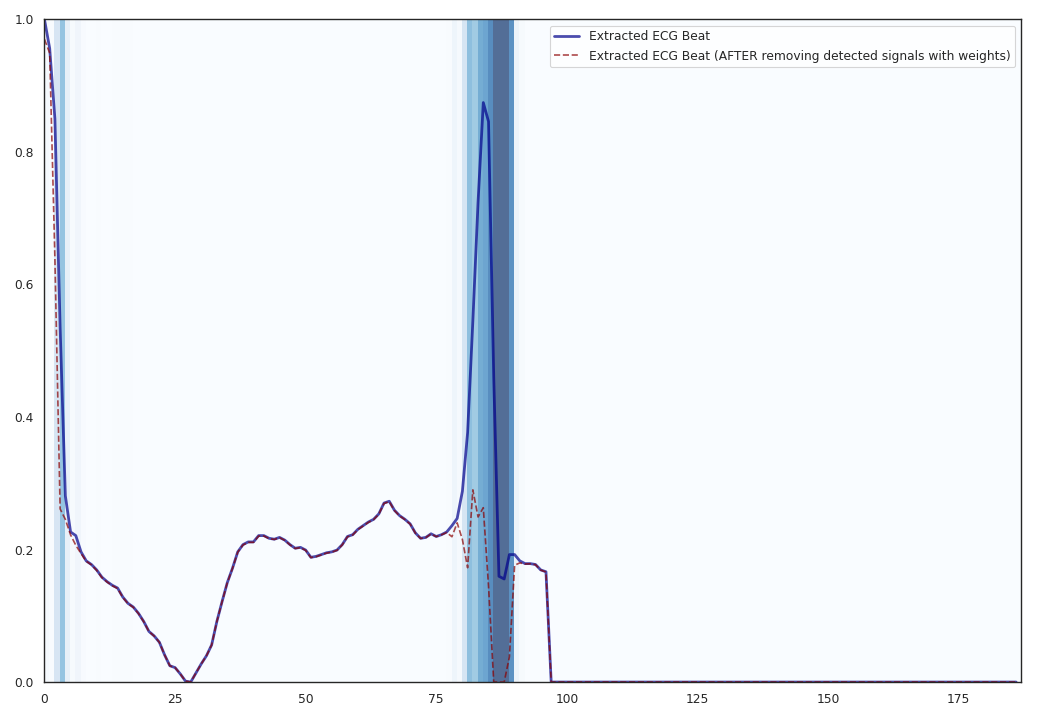}
  \includegraphics[scale=0.16]{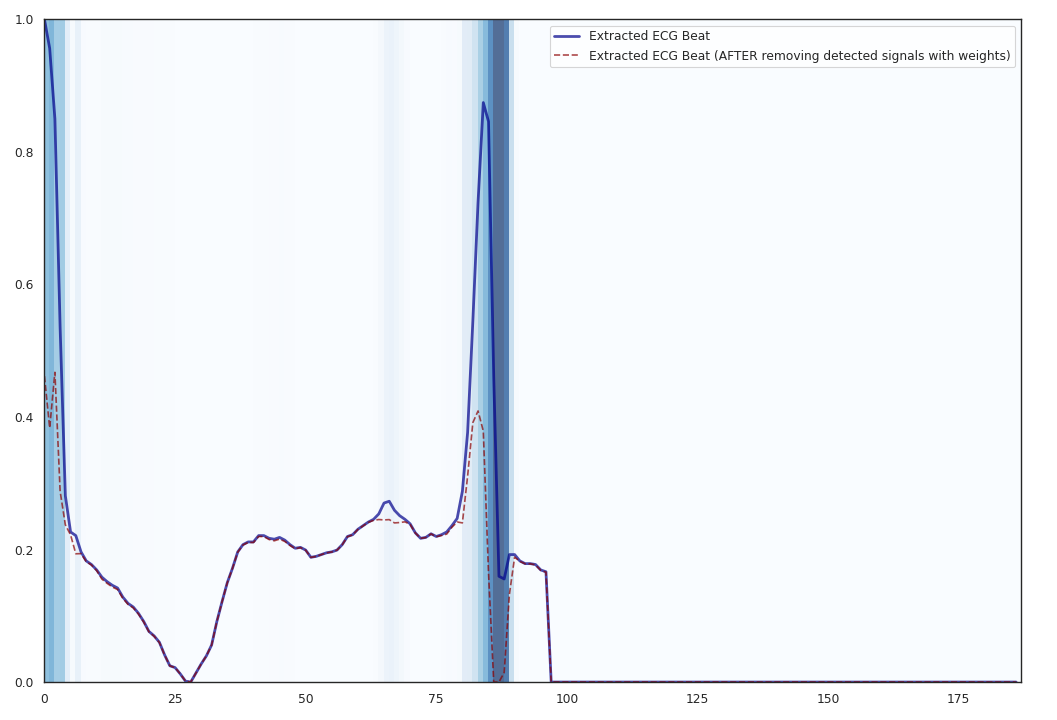}
  \includegraphics[scale=0.16]{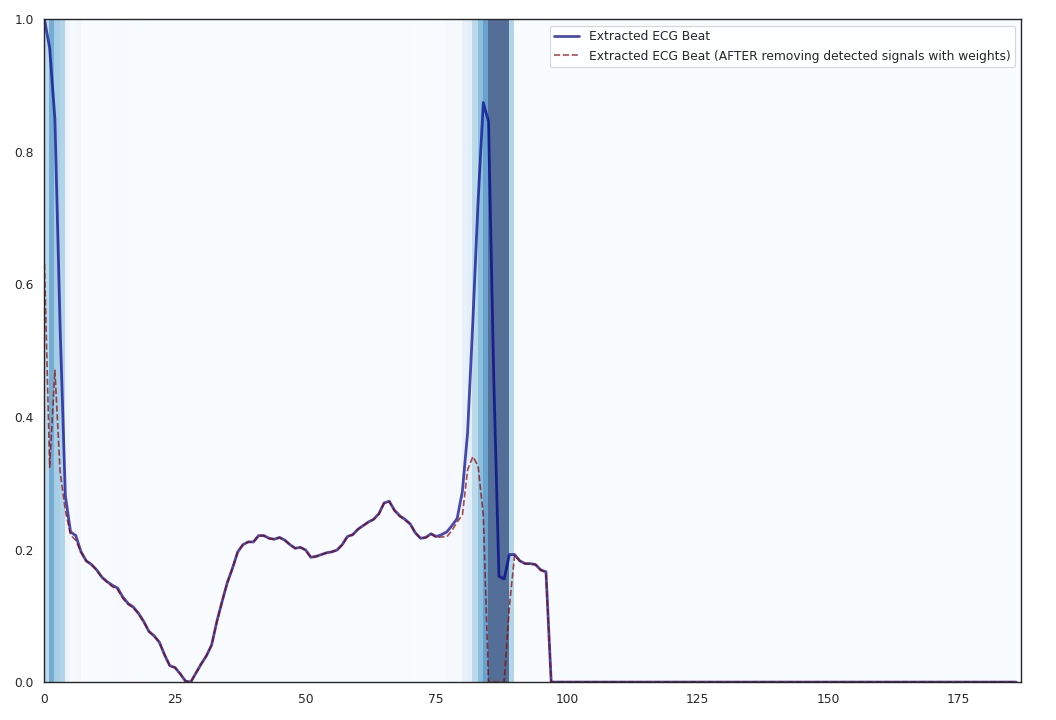}
  \caption{The localized discriminative features of one ECG signal in the MIT-BIH dataset based on the proposed framework with different CAE network architectures: \texttt{CAE64} - \texttt{CAE256}.}
  \label{fig:CAE_results}
\end{figure}

This section examines the robustness of the proposed framework against network architectures. We use the same implementation configuration with $\tau = 0.05$, and examine CAE network architectures with different numbers of neurons, denoted as \texttt{CAE64}, \texttt{CAE128}, \texttt{CAE256} and \texttt{CAE512}, where \texttt{CAE64} is constructed as: Conv1D(64)+Conv1D(32)+Conv1DTranspose(32)+Conv1DTranspose(64), and other CAE networks are defined likewise. Moreover, we also implement a localizer with a multilayer perceptron (MLP) structure: \texttt{MLP256}, \texttt{MLP512}, \texttt{MLP1024}, and \texttt{MLP2048}. For example, \texttt{MLP256} is constructed as: Dense(256)+Dense(128)+Dense(64)+Dense(187), and other MLP networks are defined likewise. As indicated in Table \ref{tab:robust}, $R^2$s of the localized discriminative features provided by convolutional auto-encoders are significantly higher and more stable than those produced by MLPs. In particular, for CAE-based networks, larger networks generally improve the performance. The localization results by the CAE networks are illustrated in Figure \ref{fig:CAE_results}: the localized discriminative features are fairly consistent with different CAE-based network architectures.

\begin{table*}[!h]\centering
  \caption{$R^2$s for the proposed framework with different network architectures. Here ``CAE'' indicates a convolutional auto-encoder, ``MLP'' indicates a multilayer perceptron, and the estimated $R^2$ is computed as in \eqref{eqn:est_r_sqare} based on 10-fold cross-validation.}
  \scalebox{1.}{
  \begin{tabular}{@{}cccccccccccccccccc@{}} \toprule
  & CAE64 & CAE128 & CAE256 & \phantom{a} & MLP512 & MLP1024 & MLP2048 \\
  & 0.816(.028) & 0.872(.011) & 0.872(.015) && 0.141(.252) & 0.156(.255) & 0.133(.242) \\
  \bottomrule
  \end{tabular}}
  \label{tab:robust}
\end{table*}

\section{Discussion}
\label{sec:conc}
XAI methods have gained prominence in many scientific domains, for example, medical diagnostics, which requires both interpretability and predictive accuracy. 
To identify discriminative features, we quantify the quality of interpretability by a generalized partial $R^2$ while measuring the interpretation effectiveness by an activity $L_1$-norm. 
On this ground, we construct a localizer by disrupting the original features, and seek a localizer yielding the most deteriorated performance of a learner while having the smallest activity norm for minimal feature disruption.
Theoretically, we show that the proposed localization method identifies discriminative features asymptotically. 
Moreover, we apply the proposed framework to the MNIST and MIT-BIH ECG datasets to interpret a learning outcome of a convolutional auto-encoder neural network. Numerical results suggest that the proposed localizer compares favorably with state-of-the-art competitors in the literature while identifying discriminative regions that are not only visually/biologically plausible but also concise. 
Furthermore, it is of interest to know if any localized features are genuinely important, for which hypothesis testing targeting a data-adaptive localizer as in \cite{dai2022significance} would be needed as a possible extension of our framework.

\begin{acks}[Acknowledgments]
  The corresponding authors for this work are Ben Dai and Wei Pan. The authors would like to thank the referees, the Associate Editor, and the Editor for the constructive feedback which greatly improved this work.
\end{acks}

\begin{funding}
  We would like to acknowledge support for this project from RGC-ECS 24302422, the CUHK direct grant, NSF DMS-1712564, DMS-1721216, DMS-1952539, and NIH grants R01GM126002, R01AG069895, R01AG065636, R01AG074858, R01AG074858, U01AG073079 and RF1 AG067924.
\end{funding}

\begin{supplement}
\stitle{Supplement to ``Data-Adaptive Discriminative Feature Localization with Statistically Guaranteed Interpretation''.} 
\sdescription{The supplementary materials consist of: Appendix A indicates that the proposed framework incorporates greedy feature selection for a linear regression model and a piecewise linear regression model; Appendix B provides details of assumptions and asymptotic results for the proposed framework; Appendix C refines the asymptotic results of the proposed framework based on a fixed $\tau$; Appendix D provides the technical proofs.}
\end{supplement}

\begin{supplement}
  \stitle{Python package dnn-locate.}
  \sdescription{The Python package
  \pkg{dnn-locate} is available in PyPi (\url{https://pypi.org/project/dnn-locate/}). For the most recent version of the package, see \url{https://github.com/statmlben/dnn-locate}.}
  \end{supplement}

  \appendix

  \section{Discriminative features for linear models}
  
  This section indicates that the proposed framework incorporates greedy feature selection for a linear regression model and a piecewise linear regression model.
  \label{sec:linear_model}
  
  \noindent \textbf{Linear regression.} Consider a linear regression model, $Y = \bm{X}^\intercal \bm{\beta}^0 + \varepsilon$, where $\mathbb{E}(\varepsilon) = 0$, random error $\varepsilon$ is independent of $\bm{X}$, and $\bm{\beta}^0$ is the true regression parameter. 
  Now, let $\bm{\delta}{(\bm{x})} = \bm{\delta} \in \mathbb{R}^p$, and $J(\bm{\delta}) = \| \bm{\delta} \|_1$, then the proposed framework (4) becomes
  \begin{equation}
  \label{eqn:model_linear_pop}
  \max_{\bm{\delta}} \ \mathbb{E} \big( \big( Y - (\bm{X} - \bm{\delta} )^\intercal \bm{\beta}^0 \big)^2 \big), \quad \text{subject to} \quad \| \bm{\delta} \|_1 \leq \tau, \quad \|\bm{\delta}\|_\infty \leq 1.
  \end{equation}
  
  
  \begin{lemma}[Greedy selection in linear regression]
  \label{lemma:sol_linear}
  A global maximizer of \eqref{eqn:model_linear_pop} is $\bm{\delta}_\tau^0(\bm{x}) = \bm{\delta}_\tau^0 = (\bm{\delta}^0_{\tau,1}, \cdots, \bm{\delta}^0_{\tau,p})^\intercal$, where
  \[\bm{\delta}^0_{\tau,j} = 
    \begin{cases} 
        \sign(\bm{\beta}^0_j), & \text{if} \quad \big| \big\{i: |\bm{\beta}^0_i| > |\bm{\beta}^0_j|, i = 1, \cdots, p \big\} \big| < \floor{\tau}, \\
        (\tau - \floor{\tau})\sign(\bm{\beta}^0_j), & \text{if} \quad \big| \big\{i: |\bm{\beta}^0_i| > |\bm{\beta}^0_j|, i = 1, \cdots, p \big\} \big| = \floor{\tau}, \\
        0, & otherwise,
     \end{cases}
  \]
  where $\floor{\tau}$ is the largest integer no greater than $\tau$. 
  \end{lemma}
  
  \noindent \textbf{Piecewise linear regression.} Next, consider a piecewise linear regression model
  \begin{equation*}
  Y = \sum_{v=1}^V \bm{X}^\intercal \bm{\beta}_v^0 \mathbb{I}( \bm{X} \in \mathcal{X}_v ) + \varepsilon, \quad \mathbb{E}(\varepsilon) = 0, 
  \end{equation*}
  where random error $\varepsilon$ is independent of $\bm{X}$, and $\mathbb{R}^p = \bigcup_{v=1}^V \mathcal{X}_v$. 
  Let $\bm{\delta}{(\bm{x})} = \bm{\delta}_v \mathbb{I}(\bm{x} \in \mathcal{X}_v)$ and $J(\bm{\delta}) = \sup_{\bm{x}} \|\bm{\delta}(\bm{x})\|_1 = \max_{v = 1, \cdots, V} \| \bm{\delta}_v \|_1$. Then the proposed framework (4) reduces to
  \begin{align}
  \label{eqn:model_piecewise_linear_pop}
  & \max_{\bm{\delta}} \ \mathbb{E} \big( \big( Y - \sum_{v=1}^V (\bm{X} - \bm{\delta}_v )^\intercal \bm{\beta}_v^0 \mathbb{I}(\bm{X} \in \mathcal{X}_v) \big)^2 \big), \nonumber \\ 
  & \text{subj to,} \ \| \bm{\delta}_v \|_1 \leq \tau, \| \bm{\delta}_v \|_\infty \leq 1, \text{ for } v = 1, \cdots, V.
  \end{align}
  
  \begin{lemma}[Greedy selection in piecewise linear regression]
  \label{lemma:sol_piecewise_linear}
  A global maximizer of \eqref{eqn:model_piecewise_linear_pop} is $\bm{\delta}_\tau^0(\bm{x}) = \bm{\delta}^0_{\tau,v} \mathbb{I}(\bm{x} \in \mathcal{X}_v)$ with $\bm{\delta}_{\tau, v}^0 = (\bm{\delta}^0_{\tau, v,1}, \cdots, 
  \bm{\delta}^0_{\tau, v,p})^\intercal$; $v = 1, \cdots, V$, where
  \[\bm{\delta}^0_{\tau, v,j} = 
    \begin{cases} 
        \sign(\bm{\beta}^0_{v,j}), & \text{if} \quad \big| \big\{i: |\bm{\beta}^0_{v,i}| > |\bm{\beta}^0_{v,j}|, i = 1, \cdots, p \big\} \big| < \floor{\tau}, \\
        (\tau - \floor{\tau})\sign(\bm{\beta}^0_{v,j}), & \text{if} \quad \big| \big\{i: |\bm{\beta}^0_{v,i}| > |\bm{\beta}^0_{v,j}|, i = 1, \cdots, p \big\} \big| = \floor{\tau}, \\
        0, & otherwise.
     \end{cases}
  \]
  \end{lemma}
  Interestingly, Lemmas \ref{lemma:sol_linear} and \ref{lemma:sol_piecewise_linear} suggest that the proposed framework produces intuitively sensible detected features for linear and piecewise linear models by taking top $\floor{\tau}$ most important features as measured by the magnitudes of the regression coefficients.

\section{Additional asymptotic results}
\label{sec:rate}

Let $\bm{\delta}^0_{\tau}$ be a global maximizer of the proposed framework over a function class
\begin{equation}
\mathcal{H}_b = \big\{ \bm{\delta} \in \mathcal{H}: \ \sup_{\bm{x}} \| \bm{\delta}(\bm{x}) \|_\infty \leq 1 \big\}.
\end{equation}
Without loss of generality, assume that $0 \leq L(d(\bm{x}_{\bm{\delta}}), Y) \leq U$ for some sufficiently large constant $U \geq 1$, for any $\bm{\delta} \in \mathcal{H}_b$ and $\bm{x} \in \mathbb{R}^p$.
Otherwise, the loss will be truncated \cite{wu2007robust}. To make the constraint sensible, we let $\tau \leq p$ since $\sup_{\bm{\delta} \in \mathcal{H}_b} J(\bm{\delta}) = p$.

Denote $\kappa_{n} = \mathbb{E} \mathcal{R}_n\big( \mathcal{H}_{b} \big)$, where $\mathcal{R}_n(\mathcal{H}_{b}) = \sup_{\bm{\delta} \in \mathcal{H}_{b}} n^{-1} \sum_{i=1}^n \big| \eta_i \big( L(d(\bm{X}_i - \bm{\delta}(\bm{X}_i)), Y) \big) \big|$ is the Rademacher complexity for $\mathcal{H}_{b}$, and $\{\eta_i\}_{i=1}^n$ are i.i.d. Rademacher random variables.

 Theorem \ref{thm:sup_asymp_r2} gives a probabilistic bound for the discrepancy between $\bm{\delta}^0_\tau$
and $\widehat{\bm{\delta}}_\tau$ in terms of the generalized $R^2$ uniformly over $\tau \in (0,p]$.

\begin{theorem}[Asymptotic $R^2$ for $\widehat{\bm{\delta}}_\tau$]
\label{thm:sup_asymp_r2}
Let $\widehat{\bm{\delta}}_\tau$ is a global maximizer of (5). For $\varepsilon_n \geq 8\kappa_{n}$ and
any learner $d$ independent of $(\bm X_i,Y_i)_{i=1}^n$, 
\begin{equation}
\mathbb{P} \Big( \sup_{\tau \in (0,p]} (R^2(d, \bm{\delta}^0_\tau) - R^2(d, \widehat{\bm{\delta}}_\tau)) \geq \varepsilon_n \Big) \leq K \exp \big( - \frac{n \varepsilon_n^2}{KU^2} \big),
\end{equation}
where $K>0$ is a constant. Hence, $ \sup_{\tau \in (0,p]} (R^2(d, \bm{\delta}^0_\tau) - R^2(d, \widehat{\bm{\delta}}_\tau)) = O_p \big( \max( \kappa_{n}, n^{-1/2} ) \big)$, where $O_p$ denotes the stochastic order.
\end{theorem}

Moreover, the asymptotics for a fixed $\tau$ is also provided in Theorem \ref{thm:asymp_r2} in Appendix \ref{sec:app_A}, where the convergence rate can be further improved. 

Next, we show that $\widehat{\bm{\delta}}_{\widehat{\tau}}$ is an asymptotically effective $r^2$-discriminative detector. Note that $\widehat{\bm{\delta}}_{\widehat{\tau}}$ already is an $r^2$-discriminative detector, since $R^2(d, \widehat{\bm{\delta}}_{\widehat{\tau}}) \geq r^2$ by the definition of $\widehat{\tau}$ in (6). Therefore, it suffices to show \textit{effectiveness}, that is, $|J(\widehat{\bm{\delta}}_{\widehat{\tau}}) - J(\bm{\delta}^0_{\tau^0})| = | \widehat{\tau} - \tau^0 | \overset{p}{\longrightarrow} 0$. To proceed, we require a smoothness condition for $R^2(d, \bm{\delta}^0_\tau)$ over $\tau$.

\noindent \textbf{Assumption A} (Smoothness).  Assume that $R^2(d, \bm{\delta}^0_\tau)$ is a continuous function in
$\tau$. Moreover, there exists a constant $\mu_0 > 0$ such that $|\tau_1 - \tau_2| \leq \mu$ if 
$| R^2(d, \bm{\delta}^0_{\tau_1}) - R^2(d, \bm{\delta}^0_{\tau_2})| \leq c_0 \mu^\alpha$ for any 
$\mu \leq \mu_0$.

\begin{theorem}[Oracle property] 
\label{thm:oracle}
Let $\bm{\delta}^0$ be an effective $r^2$-discriminative detector in Definition 2 and $\widehat{\bm{\delta}}_{\widehat{\tau}}$ be
a global maximizer of (6). Suppose Assumption A is satisfied, then for $\omega_n \geq 2(8\kappa_n / c_0)^{1/\alpha}$,
$$
\mathbb{P} \Big( \big | \widehat{\tau} - \tau^0 \big| \geq \omega_n \Big)  = \mathbb{P} \Big( \big |J(\widehat{\bm{\delta}}_{\widehat{\tau}}) - J( \bm{\delta}^0 ) \big| \geq \omega_n \Big) \leq K' \exp\Big( - \frac{n \omega^{2\alpha}}{K' U^2} \Big),
$$
where $K' > 0$ is a universal constant, which yields that $\widehat{\bm{\delta}}_{\widehat{\tau}}$ is an asymptotically effective $r^2$-discriminative detector. 
\end{theorem}

Therefore, the proposed method yields an effective $r^2$-discriminative detector as defined in (3), rendering reliable discriminative features for a target $R^2$. 

Then, we illustrate the theoretical results in Section \ref{sec:rate} to the proposed convolutional auto-encoder. For simplicity, we flatten $\bm{x}$ as a $p$-length vector, and introduce weight normalization for each kernel matrix in CAE \citep{salimans2016weight}. The detailed configuration for CAE is indicated in Table \ref{tab:CAE_conf}.
\begin{table*}[!ht]\centering
  \caption{The configuration of Encoder-CNN (E-CNN), Hidden neural network (HNN), and Decoder-CNN (D-CNN) in a convolutional auto-encoder (CAE). \texttt{WN} is a weight normalization layer, \texttt{Conv} is a convolutional layer, and \texttt{Dense} is a dense layer. }
\scalebox{1.}{
\begin{tabular}{@{}cccccccccccccccccc@{}} \toprule
 \phantom{a} & \phantom{a} & \phantom{a} & E-CNN & \phantom{a} & HNN & \phantom{a} & D-CNN & \phantom{a} \\
\midrule
& depth && $l_E$ && $l_H$ && $l_D$ \\ [.3ex]
& dim && $(p^{(i)}_E)_{i=1}^{l_E}$ && $(p^{(i)}_H)_i^{l_H}$ && $(p^{(i)}_D)_{i=1}^{l_D}$ \\ [.3ex]
& filter size && $(r^{(i)}_E)_{i=1}^{l_E}$ && -- && $(r^{(i)}_D)_{i=1}^{l_D}$ \\ [.3ex]
& \#filter && $(c^{(i)}_E)_{i=1}^{l_E}$ && -- && $(c^{(i)}_D)_{i=1}^{l_D}$ \\ [.3ex]
& basic unit && \texttt{WN}+\texttt{Conv} && \texttt{WN}+\texttt{Dense} && \texttt{WN}+\texttt{Conv} \\
\bottomrule
\end{tabular}}
\label{tab:CAE_conf}
\end{table*}

\noindent \textbf{Assumption C.} There exists a random variable $Z_i$ such that
$$
\big| L\big(d\big(\bm{X}_i - \bm{\delta}(\bm{X}_i) \big), Y_i \big) - L\big(d\big(\bm{X}_i - \bar{\bm{\delta}}(\bm{X}_i) \big), Y_i \big) \big| \leq Z_i \big\| \bm{\delta}(\bm{X}_i) - \bar{\bm{\delta}}(\bm{X}_i) \big\|_2,
$$
and $\mathbb{E}\big( \max_{i=1, \cdots, n} Z^2_i \big) \leq V_n$ for some real sequence $V_n$.

\begin{corollary}
\label{cor:rate_cnn}
Suppose that Assumptions A and C are met, and $\bm{\delta}^0 \in \mathcal{H}_b$, let $\widehat{\bm{\theta}}$ is a global maximizer of (9), and $\widehat{\bm{\delta}}_{\tau}$ is defined in (11), then
\begin{align*}
& \sup_{\tau \in (0,p]} R^2(d, \bm{\delta}^0_\tau) - R^2(d, \widehat{\bm{\delta}}_\tau) = O_p \Big( \big(\frac{ pl V_n }{n}\big)^{\frac{1}{2}} \big( \sum_{i=1}^{l_H} ( p^{(i)}_H p^{(i-1)}_H )^2 + \sum_{s \in \{E, D\}} \sum_{i=1}^{l_s} ( c_s^{(i)} r_s^{(i)} )^2 \sqrt{ \frac{p_s^{(i)}}{c_s^{(i)}} } \big)^{\frac{1}{4}} \Big), \nonumber \\
& |\widehat{\tau} - \tau^0 | = O_p \Big( \big(\frac{pl V_n }{n}\big)^{\frac{1}{2\alpha}} \big( \sum_{i=1}^{l_H} ( p^{(i)}_H p^{(i-1)}_H )^2 + \sum_{s \in \{E, D\}} \sum_{i=1}^{l_s} ( c_s^{(i)} r_s^{(i)} )^2 \sqrt{ \frac{p_s^{(i)}}{c_s^{(i)}} } \big)^{\frac{1}{4}} \big)^{\frac{1}{4\alpha}} \Big),
\end{align*}
where $l = l_E + l_H + l_D$ is the total depth of the CAE network defined in Table \ref{tab:CAE_conf}.
\end{corollary}

\section{Asymptotics for fixed \texorpdfstring{$\bm{\tau}$}{TEXT}}
\label{sec:app_A}
In this section, we establish the asymptotic properties of the proposed framework based on a fixed $\tau$. On this ground, consider the functional class:
\begin{equation}
\mathcal{H}_\tau = \big\{ \bm{\delta} \in \mathcal{H}: \sup_{\bm{x}} \| \bm{\delta}(\bm{x}) \|_\infty \leq 1, \ J(\bm{\delta}) \leq \tau \big\}.
\end{equation}

\noindent \textbf{Assumption D} (Bernstein condition). There exist constants $B \geq 1$ and $0 \leq \beta \leq 1$, such that for every $\bm{\delta} \in \mathcal{H}_\tau$, we have 
$$
\mathbb{E} \Big( \big( L\big( d(\bm{X}_{\bm{\delta}^0_\tau} ), Y\big) - L\big( d(\bm{X}_{\bm{\delta}} ), Y\big) \big)^2 \Big) \leq B \Big( \mathbb{E} \Big( L\big( d(\bm{X}_{\bm{\delta}^0_\tau} ), Y\big) - L\big( d(\bm{X}_{\bm{\delta}} ), Y\big) \Big) \Big)^\beta.
$$
This is a common condition in statistical learning theory \cite{massart2000some}, which yields that the second moment of excess risk is upper bounded by its first moment. Note that Assumption B is automatically satisfied for any distribution and functional class if $\beta = 0$. More generally, as suggested in \cite{shen2007generalization,bartlett2005local}, many regularized functional classes in learning tasks satisfy these conditions with $\beta > 0$.

\noindent \textbf{Assumption A$^\prime$} (Entropy condition). There exists $\kappa_{\tau,n} > 0$, then for any $\varepsilon \geq \kappa_{\tau,n}$, we have
$$
4 \mathbb{E} \mathcal{R}_n\big( \mathcal{H}_{\tau, \varepsilon} \big) \leq \varepsilon, \quad \mathcal{H}_{\tau, \varepsilon} = \big \{ \bm{\delta} \in \mathcal{H}_\tau: \mathbb{E} \big( L\big( d(\bm{X}_{\bm{\delta}^0_\tau} ), Y\big) - L\big( d(\bm{X}_{\bm{\delta}} ), Y\big) \big) \leq \varepsilon \big \},
$$
where $\mathcal{R}_n(\mathcal{H}_{\tau, \varepsilon}) = \sup_{\bm{\delta} \in \mathcal{H}_{\tau, \varepsilon}} n^{-1} \sum_{i=1}^n \big| \eta_i \big( L(d(\bm{X}_i - \bm{\delta}^0_{\tau}(\bm{X}_i)), Y) - L(d(\bm{X}_i - \bm{\delta}(\bm{X}_i)), Y) \big) \big|$ is the local Rademacher complexity for $\mathcal{H}_{\tau, \varepsilon}$, and $\{\eta_i\}_{i=1}^n$ are i.i.d. Rademacher random variables. 

Note that $\mathbb{E} \mathcal{R}_n\big( \mathcal{H}_{\tau, \varepsilon} \big) \leq \mathbb{E} \mathcal{R}_n\big( \mathcal{H}_{b} \big)$, hence Assumption A$^\prime$ is automatically satisfied when $\kappa_{\tau, n} \geq 4\mathbb{E} \mathcal{R}_n\big( \mathcal{H}_{b} \big)$. Moreover, when the covering number is given for the functional class, $\kappa_{\tau,n}$ can be explicitly computed.

\begin{theorem}[Asymptotics of $\bm{R}^2$]
\label{thm:asymp_r2}
Under Assumptions A$^\prime$ and D, and $\widehat{\bm{\delta}}_\tau$ is a global maximizer of (5) for a fixed $\tau > 0$, for $\varepsilon_n \geq \kappa_{\tau,n}$, then
\begin{equation}
\mathbb{P} \Big( R^2(d, \bm{\delta}^0_\tau) - R^2(d, \widehat{\bm{\delta}}_\tau) \geq \varepsilon_n \Big) \leq c_0 \exp \big( - \frac{n \varepsilon_n^{2-\beta} }{ c_1 U B } \big), 
\end{equation}
where $c_0$ and $c_1$ are positive constants, and $\kappa_{\tau,n}$ is defined in Assumption A$^\prime$. Therefore, $R^2(d, \bm{\delta}^0_\tau) - R^2(d, \widehat{\bm{\delta}}_\tau) = O_p \big( \max( \kappa_{\tau,n}, n^{-(2 - \beta)} ) \big)$.
\end{theorem}

\section{Technical proofs}

\noindent \textbf{Proof of Lemma 1.} We prove Lemma 1 by contradiction. Suppose there exists a detector $\bm{\delta}'$, such that $R^2(d, \bm{\delta}') \geq r^2$, but $J(\bm{\delta}') < J(\bm{\delta}^0_{\tau^0})$. Then, let $\tau' = J(\bm{\delta}')$, we have $R^2(d, \bm{\delta}^0_{\tau'}) \geq R^2(d, \bm{\delta}') \geq r^2,$ 
where $\bm{\delta}^0_{\tau'}$ is a global maximizer of (4) with $\tau = \tau'$. Note that $\tau' = J(\bm{\delta}') < J(\bm{\delta}^0_{\tau^0}) \leq \tau^0$, which leads to the contradiction on the definition of $\tau^0$. This completes the proof. \EOP


\noindent \textbf{Proof of Lemmas \ref{lemma:sol_linear} and \ref{lemma:sol_piecewise_linear}.} Note that $\bm{\delta}_\tau^0$ is a maximizer of 
$$
\max_{\bm{\delta}} \sum_{j=1}^p | \bm{\delta}_j \bm{\beta}_j^0 |, \quad \text{subject to,} \quad \| \bm{\delta} \|_1 \leq \tau, \quad \| \bm{\delta} \|_\infty \leq 1.
$$
Hence, $\mathbb{E} \big( \big( Y - (\bm{X} - \bm{\delta}_\tau^0 )^\intercal \bm{\beta}^0 \big)^2 \big) = |(\bm{\delta}_\tau^0)^\intercal \bm{\beta}^0| = \sum_{j=1}^p | \bm{\delta}^0_{\tau,j} \bm{\beta}_j^0 | \geq \sum_{j=1}^p | \bm{\delta}_j \bm{\beta}_j^0 | \geq |\bm{\delta}^\intercal \bm{\beta}^0| = \mathbb{E} \big( \big( Y - (\bm{X} - \bm{\delta} )^\intercal \bm{\beta}^0 \big)^2 \big)$, for any $\bm{\delta}$ satisfying the constraints. Thus, $\bm{\delta}_\tau^0$ is the global maximizer of \eqref{eqn:model_linear_pop}.

For piecewise linear regression, it suffices to consider the piecewise maximization, that is, for $v = 1, \cdots, V$, consider
$$
\max_{\bm{\delta}_v} \ \mathbb{E} \Big( \big( Y - (\bm{X} - \bm{\delta}_v )^\intercal \bm{\beta}_v^0  \big)^2 \big| \bm{X} \in \mathcal{X}_v \Big), \quad \text{subject to,} \quad \| \bm{\delta}_v \|_1 \leq C, \quad  \| \bm{\delta}_{v}\|_\infty \leq 1.
$$
Therefore, each $\bm{\delta}^0_{\tau,v}$ is provided as in Lemma \ref{lemma:sol_linear}. This completes the proof. \EOP


\noindent \textbf{Proof of Theorem \ref{thm:sup_asymp_r2}.} Let $\bm{\delta}^0_{\tau}$ and $\widehat{\bm{\delta}}_{\tau}$ be global maximizers of the population model (4) and the empirical model (5) for $\tau > 0$, then 
\begin{align}
& \sup_{\tau \in (0,p]} R^2(d, \bm{\delta}_\tau^0) - R^2(d, \widehat{\bm{\delta}}_\tau) \nonumber \\
= \ & \sup_{\tau \in (0,p]} \frac{ \mathbb{E}\big( L(d(\bm{X}), Y) \big) }{ \mathbb{E}\big( L(d(\bm{X}_{\bm{\delta}_\tau^0}), Y) \big) \mathbb{E}\big( L(d(\bm{X}_{\widehat{\bm{\delta}}_{\tau}}), Y) \big) } \Big( \mathbb{E}\big( L(d(\bm{X}_{\bm{\delta}^0_{\tau}}), Y) \big) - \mathbb{E}\big( L(d(\bm{X}_{\widehat{\bm{\delta}}_\tau}), Y) \big) \Big) \nonumber \\
\leq \ & a_0 \sup_{\tau \in (0,p]} \Big( \mathbb{E}\big( L(d(\bm{X}_{\bm{\delta}^0_{\tau}}), Y) \big) - \mathbb{E}\big( L(d(\bm{X}_{\widehat{\bm{\delta}}_\tau}), Y) \big) \Big),
\end{align}
where $a_0$ is a constant, and the last inequality follows from the fact that $\mathbb{E}\big( L(d(\bm{X}_{\bm{\delta}_\tau^0}), Y) \big)$ and $\mathbb{E}\big( L(d(\bm{X}_{\widehat{\bm{\delta}}_{\tau}}), Y) \big)$ are bounded away from zero for $\tau > 0$.

Therefore, let $l(d, \bm{\delta}) = \mathbb{E}\big( L(d(\bm{X}_{\bm{\delta}}), Y) \big)$, it suffices to consider 
\begin{align*}
& \mathbb{P} \Big( \sup_{\tau \in (0,p]} l(d, \bm{\delta}^0_{\tau}) - l(d, \widehat{\bm{\delta}}_{\tau}) \geq \varepsilon_n \Big) \\
= \ & \mathbb{P} \Big( \sup_{\tau \in (0,p]} l(d, \bm{\delta}^0_{\tau}) - L_n(d, \bm{\delta}^0_{\tau}) + L_n(d, \bm{\delta}^0_{\tau}) - L_n(d, \widehat{\bm{\delta}}_{\tau}) + L_n(d, \widehat{\bm{\delta}}_{\tau}) - l(d, \widehat{\bm{\delta}}_{\tau}) \geq \varepsilon_n \Big) \\
\leq \ & \mathbb{P} \Big( \sup_{\tau \in (0,p]} \big| l(d, \bm{\delta}^0_{\tau}) - L_n(d, \bm{\delta}^0_{\tau}) \big| + \big| L_n(d, \widehat{\bm{\delta}}_{\tau}) - l(d, \widehat{\bm{\delta}}_{\tau}) \big| \geq \varepsilon_n \Big) \\
\leq \ & \mathbb{P} \Big( \sup_{\bm{\delta} \in \mathcal{H}_b } \big| l(d, \bm{\delta}) - L_n(d, \bm{\delta}) \big| - \mathbb{E} \sup_{\bm{\delta} \in \mathcal{H}_b } \big| l(d, \bm{\delta}) - L_n(d, \bm{\delta}) \big| \geq \varepsilon_n/2 - 2\mathbb{E} \mathcal{R}_n(\mathcal{H}_b) \Big) \\
\leq \ & K \exp \big( - \frac{n \varepsilon_n^2}{KU^2} \big),
\end{align*}
where the first inequality follows from $L_n(d, \bm{\delta}^0_{\tau}) - L_n(d, \widehat{\bm{\delta}}_{\tau}) \leq 0$, the second last inequality follows from the symmetrization inequality, and the last inequality follows from Talagrand's inequality \citep{talagrand1996new,massart2000some}.

\noindent \textbf{Proof of Theorem \ref{thm:oracle}.} By Lemmas \ref{lem:emp_sol_in_bdd} and \ref{lem:pop_sol_in_bdd}, $R^2(d, \bm{\delta}^0) = r^2$ and $J(\widehat{\bm{\delta}}_{\widehat{\tau}}) = \widehat{\tau}$, thus it suffices to consider
\begin{align*}
\mathbb{P} \big( | \widehat{\tau} - \tau^0 | \geq \omega_n \big) \leq \mathbb{P} \big( \widehat{\tau} \leq \tau^0 - \omega_n \big) + \mathbb{P} \big( \widehat{\tau} \geq \tau^0 + \omega_n \big),
\end{align*}
where $\tau^0 = J(\bm{\delta}^0)$, and we treat each probability separately. Specifically,
\begin{align}
\mathbb{P} \big( \widehat{\tau} & \leq \tau^0 - \omega_n \big) \leq \mathbb{P} \big( R^2(d, \bm{\delta}^0_{\tau^0}) - R^2(d, \bm{\delta}^0_{\widehat{\tau}}) \geq c_2 \omega_n^\alpha \big) = \mathbb{P} \big( r^2 - R^2(d, \bm{\delta}^0_{\widehat{\tau}}) \geq c_2 \omega_n^\alpha \big) \nonumber \\
& \leq \mathbb{P} \big( R^2(d, \widehat{\bm{\delta}}_{\widehat{\tau}}) - R^2(d, \bm{\delta}^0_{\widehat{\tau}}) \geq c_2 \omega_n^\alpha \big) \leq \mathbb{P} \big( \sup_{\tau \in (0,p]} R^2(d, \widehat{\bm{\delta}}_{\tau}) - R^2(d, \bm{\delta}^0_{\tau}) \geq c_2 \omega_n^\alpha \big) \nonumber \\
& \leq K \exp \Big( - \frac{nc_2^2 \omega_n^{2\alpha}}{ K U^2 } \Big),
\end{align}
where the first inequality follows from Assumption A, the first equality follows from $R^2(d, \bm{\delta}^0_{\tau^0}) = r^2$, the second inequality follows from $R^2(d, \widehat{\bm{\delta}}_{\widehat{\tau}}) \geq r^2$, and the last inequality follows from Theorem \ref{thm:sup_asymp_r2}.

Next, let $\widetilde{\tau} = (\tau^0 + \widehat{\tau})/2$, then 
\begin{align*}
\mathbb{P}\big( \widehat{\tau} & \geq \tau^0 + \omega_n \big) \leq \mathbb{P}\big( R^2(d, \bm{\delta}^0_{\widetilde{\tau}}) - R^2(d, \bm{\delta}^0_{\tau^0}) \geq c_2 {(\frac{\omega_n}{2})}^\alpha \big) = \mathbb{P}\big( R^2(d, \bm{\delta}^0_{\widetilde{\tau}}) - r^2 \geq c_2 {(\frac{\omega_n}{2})}^\alpha \big)  \\
& \leq \mathbb{P}\big( R^2(d, \bm{\delta}^0_{\widetilde{\tau}}) - R^2(d, \widehat{\bm{\delta}}_{\widetilde{\tau}}) \geq c_2 {(\frac{\omega_n}{2})}^\alpha \big) \leq \mathbb{P}\big( \sup_{\tau \in (0,p] } R^2(d, \bm{\delta}^0_{{\tau}}) - R^2(d, \widehat{\bm{\delta}}_{{\tau}}) \geq c_2 {(\frac{\omega_n}{2})}^\alpha \big) \\
& \leq K \exp \Big( - \frac{nc_2^2 \omega_n^{2\alpha}}{ 2^\alpha K U^2 } \Big),
\end{align*}
where the first inequality follows from the fact that $\{ \widehat{\tau} \geq \tau^0 + \omega_n \} \subset \{ \widetilde{\tau} - \tau^0 \geq \omega_n/2 \}$, which yields that $R^2(d, \bm{\delta}^0_{\widetilde{\tau}}) - R^2(d, \bm{\delta}^0_{\tau^0}) \geq c_2 (\omega_n/2)^\alpha $ based on Assumption A. Moreover, the second last inequality follows from $R^2(d, \widehat{\bm{\delta}}_{\widetilde{\tau}}) < r^2$ by the definition of $\widehat{\tau}$, and the last inequality follows from Theorem \ref{thm:sup_asymp_r2}.

\noindent \textbf{Proof of Theorem \ref{thm:asymp_r2}.} The treatment for the proof is based on a chaining argument as in \citep{dai2019scalable,wong1995probability}.  Let $\bm{\delta}^0_{\tau}$ and $\widehat{\bm{\delta}}_{\tau}$ be global maximizers of the population model (4) and the empirical model (5) for $\tau > 0$, then 
\begin{align}
& R^2(d, \bm{\delta}_\tau^0) - R^2(d, \widehat{\bm{\delta}}_\tau) \nonumber \\
= \ & \frac{ \mathbb{E}\big( L(d(\bm{X}), Y) \big) }{ \mathbb{E}\big( L(d(\bm{X}_{\bm{\delta}_\tau^0}), Y) \big) \mathbb{E}\big( L(d(\bm{X}_{\widehat{\bm{\delta}}_{\tau}}), Y) \big) } \Big( \mathbb{E}\big( L(d(\bm{X}_{\bm{\delta}^0_{\tau}}), Y) \big) - \mathbb{E}\big( L(d(\bm{X}_{\widehat{\bm{\delta}}_\tau}), Y) \big) \Big) \nonumber \\
\leq \ & a_0 \Big( \mathbb{E}\big( L(d(\bm{X}_{\bm{\delta}^0_{\tau}}), Y) \big) - \mathbb{E}\big( L(d(\bm{X}_{\widehat{\bm{\delta}}_\tau}), Y) \big) \Big),
\end{align}
where $a_0$ is a constant, and the last inequality follows from the fact that $\mathbb{E}\big( L(d(\bm{X}_{\bm{\delta}_\tau^0}), Y) \big)$ and $\mathbb{E}\big( L(d(\bm{X}_{\widehat{\bm{\delta}}_{\tau}}), Y) \big)$ are bounded away from zero.

Therefore, it suffices to consider the asymptotic behavior of the regret error $ e(\widehat{\bm{\delta}}_\tau) = \mathbb{E}\big( L(d(\bm{X}_{\bm{\delta}^0_{\tau}}), Y) \big) - \mathbb{E}\big( L(d(\bm{X}_{\widehat{\bm{\delta}}_\tau}), Y) \big)$. 
Now, let $$A = \bigcup_{j=1}^\infty A_j, \quad A_j = \big \{ \bm{\delta} \in \mathcal{H}: \ J(\bm{\delta}) \leq \tau, \ \sup_{\bm{x}} \| \bm{\delta}(\bm{x}) \|_\infty \leq 1, \ 2^{j-1} \varepsilon_n \leq e(\bm{\delta}) \leq 2^j \varepsilon_n \big \},$$ 
we have
\begin{align}
\mathbb{P} \Big( & e( \widehat{\bm{\delta}}_\tau ) \geq \varepsilon_n \Big) \leq \mathbb{P} \Big( \sup_{ \bm{\delta} \in A} L_n(d, \bm{\delta}) - L_n(d, \bm{\delta}^0_\tau) \geq 0 \Big) \leq \sum_{j=1}^\infty \mathbb{P} \Big( \sup_{ \bm{\delta} \in A_j} L_n(d, \bm{\delta}) - L_n(d, \bm{\delta}^0_\tau) \geq 0 \Big) \nonumber \\
& \leq \sum_{j=1}^\infty \mathbb{P} \Big( \sup_{ \bm{\delta} \in A_j} \big( \mathbb{D}_n(d, \bm{\delta}) - \mathbb{D}(d, \bm{\delta}) \big) - \mathbb{E} \sup_{ \bm{\delta} \in A_j} \big( \mathbb{D}_n(d, \bm{\delta}) - \mathbb{D}(d, \bm{\delta}) \big)  \geq 2^{j-1} \varepsilon_n - 2 \mathbb{E} \mathcal{R}_j \Big) \nonumber \\
& \leq  \sum_{j=1}^\infty \mathbb{P} \Big( \sup_{ \bm{\delta} \in A_j} \big( \mathbb{D}_n(d, \bm{\delta}) - \mathbb{D}(d, \bm{\delta}) \big) - \mathbb{E} \sup_{ \bm{\delta} \in A_j} \big( \mathbb{D}_n(d, \bm{\delta}) - \mathbb{D}(d, \bm{\delta}) \big)  \geq 2^{j-2} \varepsilon_n \Big) = \sum_{j=1}^{j_0} I_j, \nonumber
\end{align}
where $\mathbb{D}_n(d, \bm{\delta}) = L_n(d, \bm{\delta}) - L_n(d, \bm{\delta}^0_\tau)$, $\mathbb{D}(d, \bm{\delta}) = \mathbb{E}\big(\mathbb{D}_n(d, \bm{\delta})\big)$, the last equality follows from $I_j = 0$ if $j > j_0$ with $j_0 = \max_{j} \{ j: 2^{j-2} \varepsilon_n \leq 2U_r \}$, and the second last inequality follows from the symmetrization inequality \citep{koltchinskii2011oracle} and Assumption A,
$$
\mathbb{E} \sup_{ \bm{\delta} \in A_j} \big( \mathbb{D}_n(d, \bm{\delta}) - \mathbb{D}(d, \bm{\delta}) \big) \leq 2 \mathbb{E} \mathcal{R}_j \leq 2^{j-2} \varepsilon_n,
$$
where $\mathcal{R}_j = \sup_{\bm{\delta} \in A_j} n^{-1} \big| \sum_{i=1}^n \eta_i \big( L(d(\bm{X}_i - \bm{\delta}^0_{\tau}(\bm{X}_i)), Y) - L(d(\bm{X}_i - \bm{\delta}(\bm{X}_i)), Y) \big) \big|$ is the local Rademacher complexity for $A_j$, and $\{\eta_i\}_{i=1}^n$ are i.i.d. Rademacher random variables. Therefore, it suffices to bound each $I_j$ separately. By Talagrand's concentration inequalities \citep{talagrand1996new} and the symmetrization inequality \citep{koltchinskii2011oracle}, for $j = 1, \cdots, j_0$, we have 
\begin{align}
I_j & \leq K \exp \big( - \frac{n (2^{j} \varepsilon_n)^2 }{32a_1 (\sigma_j^2 + \mathbb{E} \mathcal{R}_j + 2^{j-1} \varepsilon_n  ) } \big) \leq K \exp \big( - \frac{n (2^{j} \varepsilon_n)^2 }{32a_1 ( B (2^j \varepsilon_n)^\beta + 2^{j-3} \varepsilon_n + 2^{j-1} \varepsilon_n  ) } \big) \nonumber \\
& \leq K \exp \big( - \frac{n (2^{j-3} \varepsilon_n / U)^2 }{4a_1/U ( B (2^{j-3} \varepsilon_n / U)^\beta + B 2^{j-3} \varepsilon_n / U  ) } \big) \leq K \exp \big( - \frac{n (2^{j} \varepsilon_n)^{2-\beta} }{512 a_1 B U  } \big),
\end{align}
where $K$ is a universal constant, $\sigma_j^2 = \sup_{\bm{\delta} \in A_j} \mathbb{E} \big( \big( L(d(\bm{X}_{\bm{\delta}^0_{\tau}}), Y) - L(d(\bm{X}_{\bm{\delta}}), Y) \big)^2 \big)$, and the last inequality follows from the fact that $(2^{j-4} \varepsilon_n / U)^\beta \geq (2^{j-4} \varepsilon_n / U) $, since $2^{j-2} \varepsilon_n \leq 2 U_r$. Therefore, 
\begin{align*}
\mathbb{P} \Big( e(\widehat{\bm{\delta}}_\tau) \geq \varepsilon_n \Big) & \leq \sum_{j=1}^{j_0} K \exp \big( - \frac{n (2^{j} \varepsilon_n)^{2-\beta} }{512 a_1 B U  } \big) \\ 
& \leq \sum_{j=1}^{\infty} K \exp \big( - \frac{n (2^{j} \varepsilon_n)^{2-\beta} }{512 a_1 B U  } \big) \leq c_0 \exp \big( - \frac{n \varepsilon_n^{2-\beta} }{256 a_1 B U  } \big).
\end{align*}
This completes the proof. \EOP

\begin{lemma}
\label{lem:emp_sol_in_bdd}
Let $\widehat{\bm{\delta}}_{\widehat{\tau}}$ be a global maximizer of (5), and $\widehat{\tau} = \min \{ \tau \in (0,p]: R^2(d, \widehat{\bm{\delta}}_{{\tau}}) \geq r^2 \}$, then $J(\widehat{\bm{\delta}}_{\widehat{\tau}}) = \widehat{\tau}$.
\end{lemma}

\noindent \textbf{Proof of Lemma \ref{lem:emp_sol_in_bdd}.} We prove Lemma \ref{lem:emp_sol_in_bdd} by contradiction. Suppose $J(\widehat{\bm{\delta}}_{\widehat{\tau}}) < \widehat{\tau}$, and $\widehat{\bm{\delta}}_{\widehat{\tau}}$ is a global maximizer of $L_n(d, \bm{\delta})$ in (5). Now, let $\tau' = J(\widehat{\bm{\delta}}_{\widehat{\tau}}) < \widehat{\tau}$, then $\widehat{\bm{\delta}}_{\tau'} = \widehat{\bm{\delta}}_{\widehat{\tau}}$ is a global maximizer of $L_n(d, \bm{\delta})$ in (5). Thus, $R^2(d, \widehat{\bm{\delta}}_{\tau'}) = R^2(d, \widehat{\bm{\delta}}_{\widehat{\tau}}) \geq r^2$, and $\tau' < \widehat{\tau}$, which contradicts to the definition of $\widehat{\tau}$. This completes the proof. \EOP

\begin{lemma}
\label{lem:pop_sol_in_bdd}
Under Assumption A, let $\bm{\delta}^0_{\tau}$ as a global maximizer of (4), then for any $0\leq r^2 \leq r^2_{\max}$, there exists $\tau_0$, such that $R^2(d, \bm{\delta}^0_{\tau_0}) = r^2$. Moreover, the solution is achieved at the boundary, that is, $\tau = J(\bm{\delta}^0_{\tau})$.
\end{lemma}

\noindent \textbf{Proof of Lemma \ref{lem:pop_sol_in_bdd}.} By Assumption A, $R^2(d, \bm{\delta}^0_\tau)$ is a continuous function with respect to $\tau$, then for any $0\leq r^2 \leq r^2_{\max}$, there exists $\tau'$, such that $R^2(d, \bm{\delta}^0_{\tau'}) = r^2$. Next, we prove that the solution is achieved at the boundary by contradiction. Suppose there exists $\tau > 0$, such that $\tau' = J(\bm{\delta}^0_{\tau}) < \tau$, then $\bm{\delta}^0_{\tau'} = \bm{\delta}^0_{\tau}$. According to Assumption A, we have $\tau' = \tau$, since $| R^2(d, \bm{\delta}^0_{\tau}) - R^2(d, \bm{\delta}^0_{\tau'}) | = 0$, which leads to the contradiction to the fact that $\tau' < \tau$. This completes the proof. \EOP

\noindent \textbf{Proof of Corollary \ref{cor:rate_cnn}.} Based on Theorems \ref{thm:sup_asymp_r2} and \ref{thm:oracle}, it suffices to compute the Rademacher complexities for the function space in (8). For any $\bm{\delta}_\tau$, $\bar{\bm{\delta}}_\tau \in \mathcal{H}_b^\tau$, 
\begin{align}
& n^{-1} \sum_{i=1}^n | L\big(d\big(\bm{x}_i - \bm{\delta}_\tau(\bm{x}_i), y_i \big) - L\big(d\big(\bm{x}_i - \bar{\bm{\delta}}_\tau(\bm{x}_i), y_i \big)|^2 \leq n^{-1} \sum_{i=1}^n z_i^2 \big\| \bm{\delta}_\tau(\bm{x}_i) - \bar{\bm{\delta}}_\tau(\bm{x}_i) \big\|^2_2 \nonumber \\
& \leq n^{-1} \tau^2 p^2 \sum_{i=1}^n z_i^2 \big\| \CAE_{\bm{\theta}}(\bm{x}_i) - \CAE_{\bar{\bm{\theta}}}(\bm{x}_i) \big\|^2_2 \leq n^{-1} \tau^2 p^2 \| \bm{z}^2 \|^2_n  \sum_{i=1}^n \big\| \CAE_{\bm{\theta}}(\bm{x}_i) - \CAE_{\bar{\bm{\theta}}}(\bm{x}_i) \big\|^2_2, \nonumber
\end{align}
where $\|\bm{z}^2\|^2_n = \max_{i=1,\cdots,n} z_i^2 $, the first inequality follows from Assumption D, and the second inequality follows from Lipschitz conditions for the truncated ReLU and softmax functions. Hence, the entropy for $\mathcal{L} = \big\{ L( d(\bm{X} - \bm{\delta}(\bm{X})), Y): \bm{\delta} \in \mathcal{H}_b^\tau \big\}$ is bounded by
\begin{align}
\log \mathcal{N} \big( \mathcal{L}, u \big) & \leq \log \mathcal{N} \big( \mathcal{H}_{\CAE}, \frac{u}{\tau p \| \bm{z}^2 \|_n} \big) \leq \Big(\frac{2 p^2 \|\bm{z}^2\|_n }{u}  ( l_E + l_H + l_D )^2 \big( \sum_{i=1}^{l_H} ( p^{(i)}_H p^{(i-1)}_H )^2 \nonumber \\
& \quad + \sum_{i=1}^{l_E} ( c_E^{(i)} r_E^{(i)} )^2 \sqrt{ p_E^{(i)} / c_E^{(i)}} + \sum_{i=1}^{l_D} ( c_D^{(i)} r_D^{(i)} )^2 \sqrt{ p_D^{(i)} / c_D^{(i)}} \big)  \Big)^{\frac{1}{2}},
\end{align}
The second inequality follows from Lemma 14 in \cite{lin2019generalization} and $\tau \leq p$, where the covering number for the functional space $\mathcal{H}_{\CAE} = \big \{ \CAE_{\bm{\theta}}(\bm{x}) | \bm{\theta} \in \bm{\Theta} \big \}$ is upper bounded. Note that all weight matrices in CAE are normalized by `WN' layer, thus both spectral norm and Frobenius norm are upper bounded by one. By Theorem 3.12 of \cite{koltchinskii2011oracle}, the Rademacher complexity is upper bounded by
\begin{align*}
\mathbb{E} \mathcal{R}_n(\mathcal{L}) & \leq a_0 n^{-1/2}\Big(2 p^2 \mathbb{E} \|\bm{z}^2\|_n ( l_E + l_H + l_D )^2 \big( \sum_{i=1}^{l_H} ( p^{(i)}_H p^{(i-1)}_H )^2 \nonumber \\
&  + \sum_{i=1}^{l_E} ( c_E^{(i)} r_E^{(i)} )^2 \sqrt{ p_E^{(i)} / c_E^{(i)}} + \sum_{i=1}^{l_D} ( c_D^{(i)} r_D^{(i)} )^2 \sqrt{ p_D^{(i)} / c_D^{(i)}} \big)  \Big)^{\frac{1}{4}}.
\end{align*}
Therefore, the desirable results follow from Theorem \ref{thm:oracle}. \EOP 


\bibliographystyle{imsart-nameyear} 
\bibliography{detect}       





\end{document}